\documentclass[sigconf]{acmart}
\usepackage{subcaption}
\usepackage{xcolor}
\usepackage[table]{xcolor}
\AtBeginDocument{%
  }

\copyrightyear{2025}
\acmYear{2025}
\setcopyright{acmlicensed}\acmConference[MM '25]{Proceedings of the 33rd ACM International Conference on Multimedia}{October 27--31, 2025}{Dublin, Ireland}
\acmBooktitle{Proceedings of the 33rd ACM International Conference on Multimedia (MM '25), October 27--31, 2025, Dublin, Ireland}
\acmDOI{10.1145/3746027.3754941}
\acmISBN{979-8-4007-2035-2/2025/10}

\begin{document}

\title{Graph Unlearning Meets Influence-aware Negative \\
Preference Optimization}

\author{Qiang Chen}
\authornote{Both authors contributed equally to this research.}
\email{qiangchen.sh@gmail.com}
\orcid{0009-0000-4336-2392}
\affiliation{%
  \institution{Central South University}
  \city{Changsha}
  \country{China}
}

\author{Zhongze Wu}
\authornotemark[1]
\email{wzz0413@csu.edu.cn}
\affiliation{%
  \institution{Central South University}
  \city{Changsha}
  \country{China}
}

\author{Ang He}
\email{heang@stu.shmtu.edu.cn}
\affiliation{%
  \institution{Shanghai Maritime University}
  \city{Shanghai}
  \country{China}
}

\author{Xi Lin}
\authornote{Both authors are corresponding authors.}
\email{linxi234@sjtu.edu.cn}
\affiliation{%
  \institution{Shanghai Jiao Tong University}
  \city{Shanghai}
  \country{China}
}

\author{Shuo Jiang}
\email{jiangshuo@tongji.edu.cn}
\affiliation{%
 \institution{Tongji University}
 \city{Shanghai}
 \country{China}}

\author{Shan You}
\email{youshan@senseauto.com}
\affiliation{%
  \institution{SenseTime Research}
  \city{Beijing}
  \country{China}}

\author{Chang Xu}
\email{c.xu@sydney.edu.au}
\affiliation{%
  \institution{University of Sydney}
  \city{Sydney}
  \country{Australia}}

\author{Yi Chen}
\email{yichen@ust.hk}
\affiliation{%
  \institution{Hong Kong University of Science and Technology}
  \city{Hong Kong}
  \country{China}}

\author{Xiu Su}
\authornotemark[2]
\email{xiusu1994@csu.edu.cn}
\affiliation{%
  \institution{Central South University}
  \city{Changsha}
  \country{China}
}

\renewcommand{\shortauthors}{Qiang Chen et al.}

\begin{abstract}
Recent advancements in graph unlearning models have enhanced model utility by preserving the node representation essentially invariant, while using gradient ascent on the forget set to achieve unlearning. However, this approach causes a drastic degradation in model utility during the unlearning process due to the rapid divergence speed of gradient ascent. In this paper, we introduce \textbf{INPO}, an \textbf{I}nfluence-aware \textbf{N}egative \textbf{P}reference \textbf{O}ptimization framework that focuses on slowing the divergence speed and improving the robustness of the model utility to the unlearning process. Specifically, we first analyze that NPO has slower divergence speed and theoretically propose that unlearning high-influence edges can reduce impact of unlearning. We design an influence-aware message function to amplify the influence of unlearned edges and mitigate the tight topological coupling between the forget set and the retain set. The influence of each edge is quickly estimated by a removal-based method. Additionally, we propose a topological entropy loss from the perspective of topology to avoid excessive information loss in the local structure during unlearning. Extensive experiments conducted on five real-world datasets demonstrate that INPO-based model achieves state-of-the-art performance on all forget quality metrics while maintaining the model's utility. Codes are available at \href{https://github.com/sh-qiangchen/INPO}{https://github.com/sh-qiangchen/INPO}. 
\end{abstract}




\maketitle

\section{Introduction}
Graph-structured data\cite{Liang2024:56,Liang2024,Li2024,Shi2024} play a pivotal role in multimodal models, facilitating the discovery of relevant information among entities. To better capture the relationships, Graph Neural Networks (GNNs)\cite{Kipf, Hamilton} have recently emerged as a crucial tool. With increasing awareness of privacy protection and the introduction of regulatory policies\cite{Voigt2017,CCPA2018}, removing some privacy-related information from trained graph models is urgent. This has motivated a line of research on graph unlearning, aiming to strengthen \textit{the Right to be Forgotten}. Moreover, graph unlearning is also highly valuable for removing inaccurate or outdated information contained in training samples. 

Graph unlearning\cite{Cheng2023} refers to the process of forgetting or removing information related to certain features, edges and nodes from a pre-trained graph model. Designing graph unlearning models is challenging due to the strong coupling relationships between elements in graph data. Currently, most models\cite{Wu2023GIF,Liu2022TheRight, Tan2024Unlink,Li2024Towards} rely on distance-based loss to preserve the predictive performance of the model on the retention set, while effectively forgetting using gradient ascent. Specially, GNNDelete\cite{Cheng2023} facilitates edge unlearning by minimizing the MSE loss between the embeddings of deleted edges and those that were non-existent, and makes it infeasible to distinguish the representation distance between the forgot set and the retain data. Meanwhile, the model\cite{Li2024Towards} based on gradient ascent exhibits \textbf{rapid divergence speed}, \textbf{ significantly degrading model utility as unlearning progresses}.

\begin{figure}[t]
    \centering
    \begin{subfigure}[b]{0.23\textwidth}
        \includegraphics[width=\textwidth]{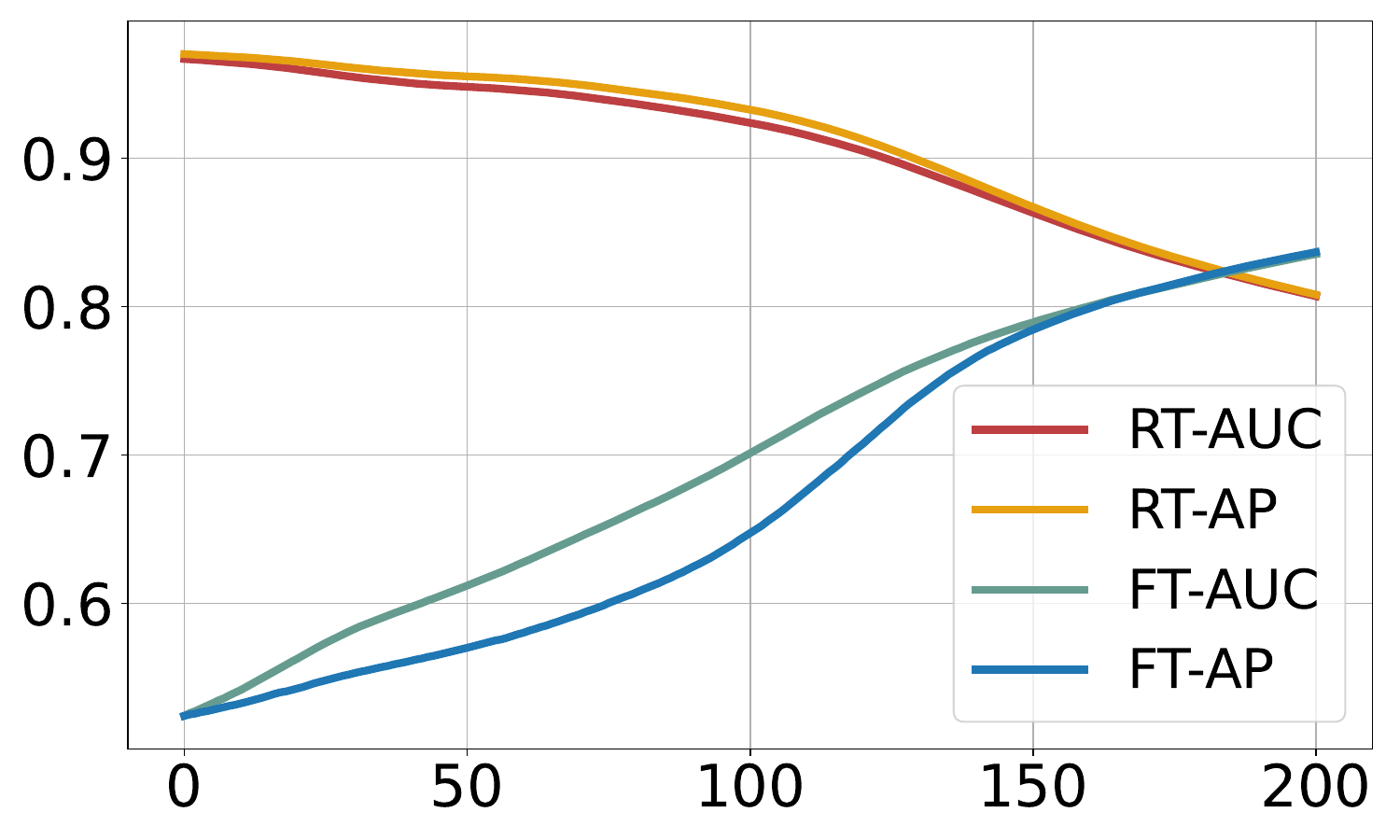}
        \caption{Accuracy Changes}
        \label{fig1:left}
    \end{subfigure}
    \hfill
    \begin{subfigure}[b]{0.23\textwidth}
        \includegraphics[width=\textwidth]{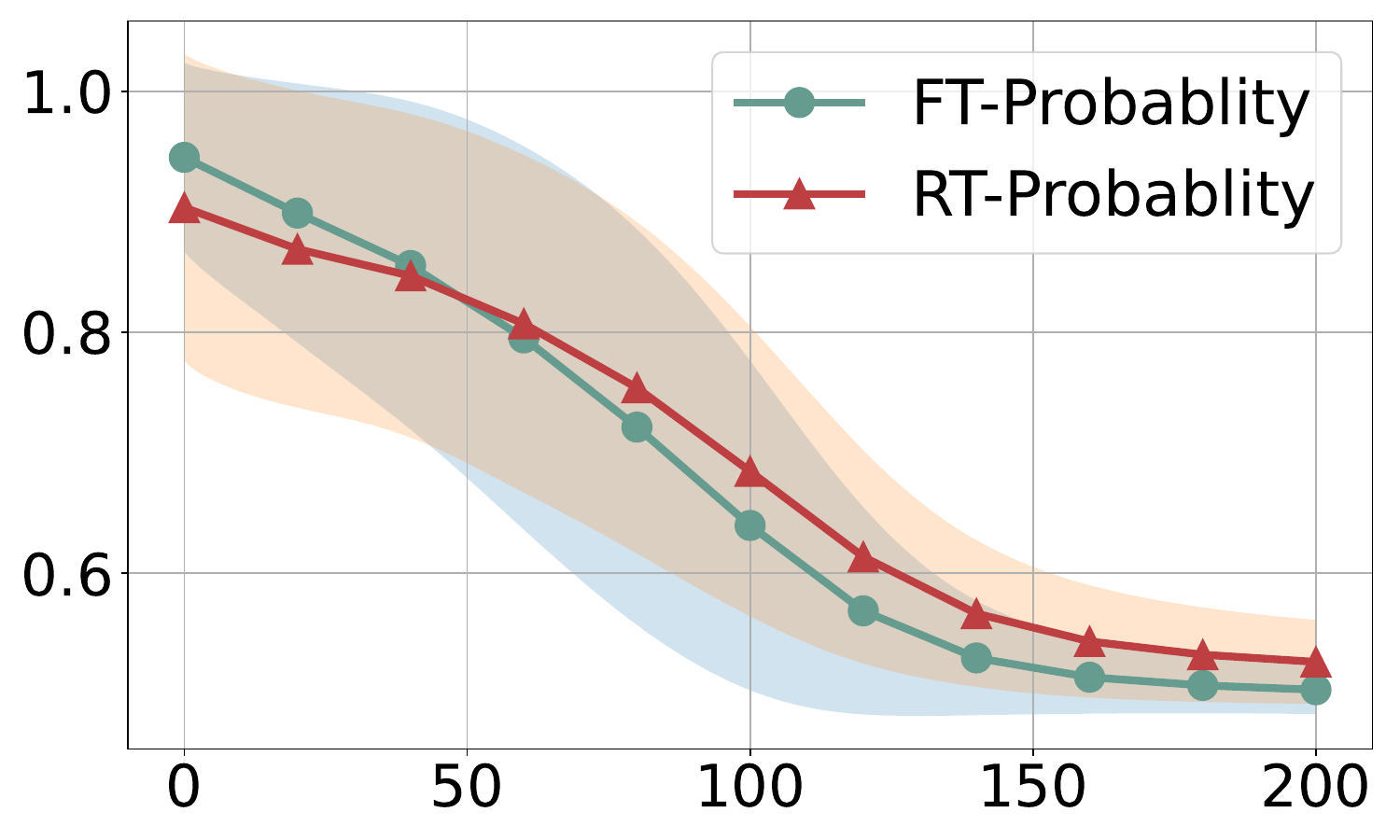}
        \caption{Probability Changes}
        \label{fig1:right}
    \end{subfigure}
    \caption{The accuracy and probability curve using NPO on RT and FT of DBLP. RT and FT denote the retain set and the forget set, respectively. }
    \label{fig1:both}
\end{figure} 

Recently, Reinforcement Learning from Human Feedback (RLHF) offers a preference optimization manner\cite{Li2023, Li2023Aaron,Zhao2025DO,Li2025LEARNING} to learn value alignment, and its superior performance has been demonstrated in crucial tasks such as LLM Unlearning and LLM Safety\cite{Dai2024, Ji2024, Bai2025ONLINE, Cha2025TOWARDS, Wu2025TOWARDS}. Direct Preference Optimization(DPO)\cite{Rafailov2023} derives a straightforward approach for policy optimization by directly using preferences, thus avoiding the complexity of learning a reward function. Negative Preference Optimization(NPO)\cite{Zhang2024} ignores the positive samples used in DPO and optimizes using only negative samples, achieving a better balance between model forget quality and utility. Furthermore, NPO-based and DPO-based methods\cite{Yuan2025,Wang2025,Scholten2025PROBABILISTIC,Wang2025RETHINKING} have shown excellent performance in LLM Unlearning tasks due to \textbf{slower divergence speed}, reducing the impact on model utility when executing an unlearning goal. Hence, a natural question arises: "\textbf{Are preference optimization method effective in graph unlearning tasks where data entities are strongly couple}?"

To explore this, we conduct a pilot study to investigate the impact of graph unlearning on model utility. As shown in Figure \ref{fig1:left}, as the AUC and AP on the forget set improve, their counterparts on the retain set exhibit a corresponding decreases. Figure \ref{fig1:right} indicates enhancing the model's ability to forget specific data instances leads to a decrement in prediction probability over previously learned data, reflecting the challenge in balancing model forget quality and utility. These two phenomena indicate that \textbf{the robustness of the model utility to the graph unlearning is insufficient}.

\begin{figure*}[t]
  \centering
    \includegraphics[width=1\linewidth]{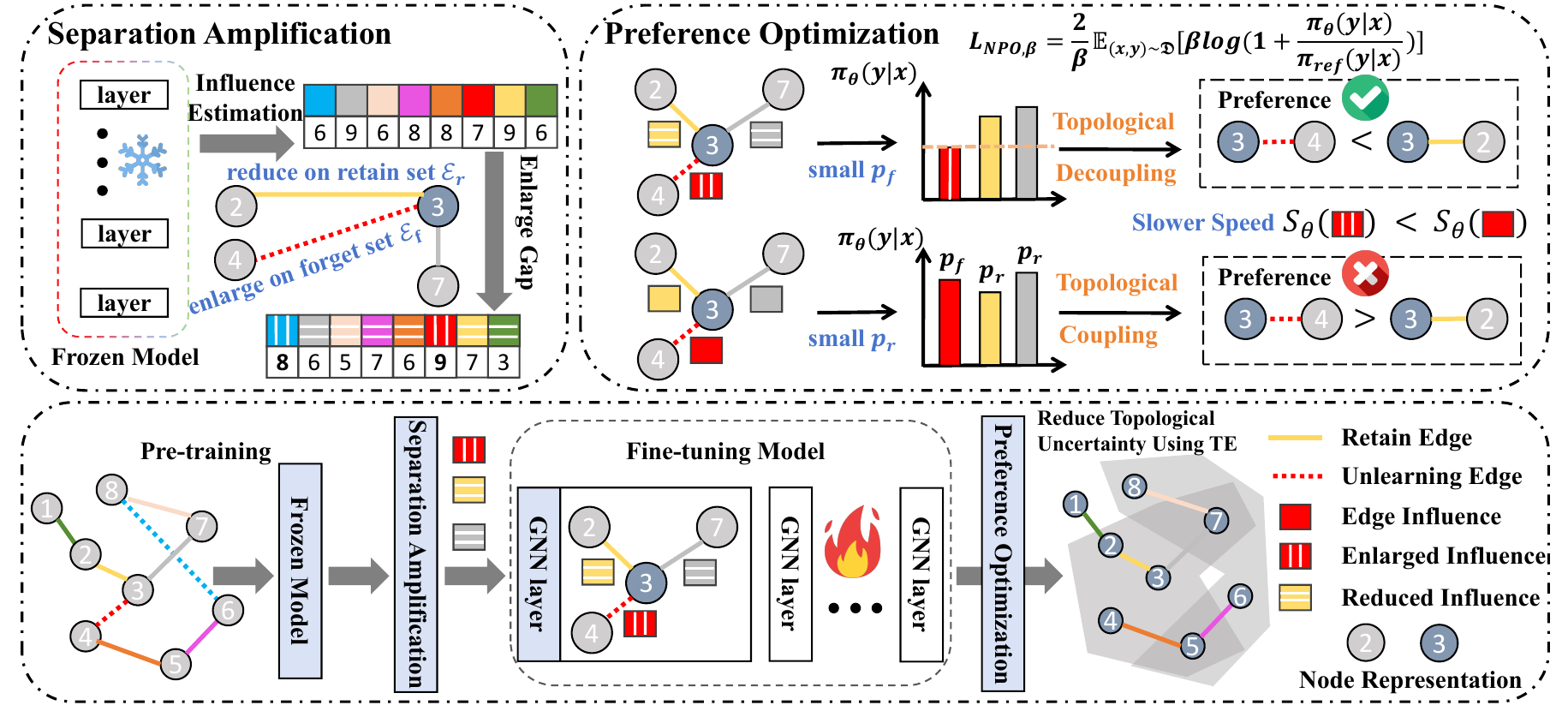}
  \caption{Overview of our INPO. Achieving graph unlearning by preference optimization considering topological decoupling.}
  \label{fig:framework}
\end{figure*}

In this work, we propose an Influence-aware Negative Preference Optimization framework to mitigate the tight topological coupling between the forget set and the retain set, as shown in Figure \ref{fig:framework}, aiming to improve the robustness of the model utility to the unlearning process. Specifically, we first analyze the small adaptive coefficient of NPO is beneficial for the robustness and theoretically propose that unlearning high-influence edges can reduce impact on the retain set to improve the robustness, which is achieved by \textbf{enlarging the probability difference between the forget set and the retain set}. Based on this insight, we develop an influence-aware message function to amplify the influence of unlearned edges and mitigate the tight topological coupling. Our message function incorporates the influence of edges into GNN, and the influence of each edge is quickly estimated by a removal-based method. This method is fast, requiring only a single inference, and does not impose any additional computational overhead. The proposed new message function achieves a result similar to forgetting high-impact edges. Additionally, to further preserve effective model utility, we propose a topological entropy loss function from the perspective of topology to avoid excessive information loss in the local structure before and after unlearning.  

In summary, the main contributions of our paper are:
\begin{itemize}
\item We are the first to propose a preference optimization approach to improve the robustness of the model utility to graph unlearning. 
\item We theoretically propose that unlearning high-influence edges can improve the robustness and design a novel message function to amplify the effects of unlearned edges to improve the robustness. 
\item We propose a topological entropy loss function from the perspective of topology to avoid excessive information loss in the local structure before and after unlearning. 
\item We validated the effectiveness of \textbf{INPO} on five real-world datasets. The experimental results strongly indicate that INPO achieves state-of-the-art performance on all forgetting quality metrics while maintaining the model's utility. On the DBLP and Cora datasets, the performance of MI Ration improved by \(6.5\%\) and \(2.3\%\), respectively.
\end{itemize}

\section{Preliminaries}

\subsection{Graph Unlearning}
Graph unlearning tasks consist mainly of three types: feature unlearning, node unlearning, and edge unlearning. This work focuses primarily on edge unlearning. Given a pre-trained model (i.e., the reference model) parameterized by \(\theta_{ref}\), an attributed graph \(G=(\mathcal{V},\mathcal{E},\textbf{X})\) with \(N = |\mathcal{V}|\) nodes, set of edges \(\mathcal{E}=\{(v_i, v_j)\}_{i,j=1}^{N}\), and d-dimensional node features \(\textbf{X} = \{\textbf{x}_0, \ldots, \textbf{x}_{N-1}\}\) where \(\textbf{x}_i \in \mathbb{R}^{d}\) is used as dataset. Edge unlearning requires fine-tuning the pre-trained model \cite{su2022searching, su2021bcnet, su2021prioritized} to forget some edges (i.e., the forget set) \(\mathcal{E}_f \subseteq \mathcal{E}\) that are specified by a deletion request, while preserving performance on the retain set \(\mathcal{E}_r = \mathcal{E} - \mathcal{E}_f\). In other words, we would like the unlearned model to behave as if the edges in forget set were never used to train.

\subsection{Graph Neural Network}
Modern GNN follow the message passing mechanism, which iteratively updates the representation of a node by aggregating representations of its neighbors. Formally, the update of node \(v \in \mathcal{V}\) at GNN's \textit{i}-th layer can be expressed by:  
\begin{equation}
\label{e1}
h_v^{(i)} = ReLu(w_0^{(i)}h_v^{(i-1)} + w_1^{(i)}\sum\limits_{u \in N(v)} \rho_{v, u} h_u^{(i-1)}),
\end{equation}
where \(h_v^{(i)} \in \mathbb{R}^{d_i}\) is the embedding vector of node \(v\) at the \textit{i}-th layer, \(w_0^{(i)}\) and \(w_1^{(i)}\) are weight matrices in \(\mathbb{R}^{d_i \times d_{i-1}}\). The activation function for all layers is Relu \cite{su2021k, su2022vitas, su2021locally}, except for the last layer. \(N(v)\) represents the 1-hop neighbors of node \(v\), and \(\rho\) is the normalized weight between two nodes.

\subsection{Negative Preference Optimization}


In preference optimization, only a negative response \(y_l\) is provided, the NPO loss is calculated without any positive response. Specifically, it is: 
\begin{equation}
\begin{aligned}
\label{eq:npo}
\mathcal{L}_{NPO, \beta}(\theta) = \frac{2}{\beta} \mathbb{E}_{(x, y_l) \sim \mathcal{D}} [log (1 + \frac{\pi_\theta(y_l \mid x)}{\pi_{ref}(y_l \mid x)} )^{\beta} ],
\end{aligned}
\end{equation}
Minimizing \(\mathcal{L}_{NPO, \beta}\) guarantees that the prediction probability of \(y_l\) is as small as possible.

\section{Robustness Against Unlearning}
To the best of our knowledge, we first propose that edge unlearning can be viewed as a preference optimization problem. We treat the prediction of each edge \((\mathcal{V}_i, \mathcal{V}_j) \in \mathcal{E}_f\) as a negative response, and use NPO loss to optimize. The prediction probability of pre-trained model on the forget set is directly used as the reference policy \(\pi_{ref}\). Minimizing \(\mathcal{L}_{NPO, \beta}\) ensures that the prediction probability of each edge \((\mathcal{V}_i, \mathcal{V}_j) \in \mathcal{E}_f\) is as small as possible, aligning with the goal of unlearning edges in the forget set.

NPO\cite{Zhang2024, Fan2024Simplicity} indicates that the decrease of model utility is positively correlated with the model's divergence speed during unlearning, which corresponds to the gradient of the NPO loss. The gradients are as follows:
\begin{equation}
\label{npo:gradient}
\nabla_\theta \mathcal{L}_{\text{NPO},\beta} = \mathbb{E}_{\mathcal{E}_f} \left[ S_\theta(x, y) \nabla_\theta \log \pi_\theta(y \mid x) \right],
\end{equation}
where \(S_\theta(x, y)=2 \pi_\theta^\beta(y \mid x) / [ \pi_\theta^\beta(y \mid x) + \pi_{ref}^\beta(y \mid x) ]\) can be views as an adaptive coefficient.

\begin{lemma}
\label{lemma:3.1}
If the predicted probability of unlearned model much less than the counterpart of original pre-trained model on the forget set, i.e., \(\pi_\theta(y \mid x) \ll \pi_{ref}(y \mid x)\), the performance decrease on the retain set is slow.
\end{lemma}

According to equation \ref{npo:gradient}, we know that \(S_\theta(x, y) \ll 1\) when \(\pi_\theta(y \mid x) \ll \pi_{ref}(y \mid x)\), which means NPO diverge much slower than GA loss(\(\nabla_\theta \mathcal{L}_{\text{GA}} = \mathbb{E}_{\mathcal{E}_f} \left[\nabla_\theta \log \pi_\theta(y \mid x) \right]\)).However, as the probability of forgetting edges decreases, \textbf{the probability of retaining edges also decreases due to topological coupling in graph unlearning task}.  

\begin{figure}[t]
  \centering
  \includegraphics[width=0.7\linewidth]{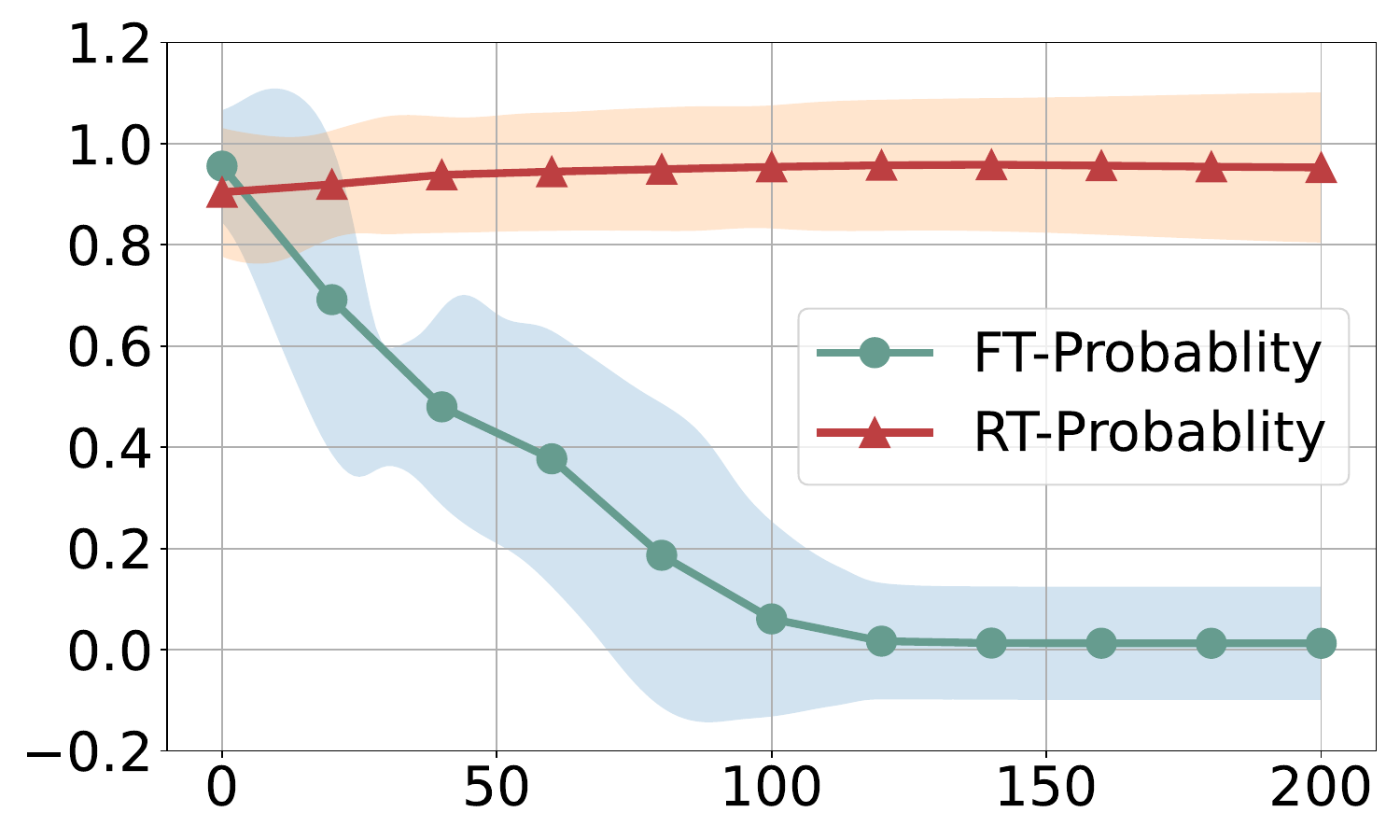}
  \caption{The predicted probability after unlearning high-influence edges on DBLP.}
  \label{fig2}
\end{figure}

A considerable corpus of research\cite{Wang2025TOWARDS,Tang2020investigating,Liu2021tail,Chen2024link,fan2025manta,zhu2025interpretable} has substantiated that nodes or edges with high influence lead to better performance in link prediction tasks, i.e., the predicted probability for the existence of unlearned edge is relatively low compared to an counterpart on the retain set as unlearning progresses. Therefore, amplifying the influence of unlearned edges to \textbf{enlarge the probability difference}(we give theoretical proof in Supplementary Materials A.4) between the forget set and the retain set can mitigate topological coupling, i.e., \(\pi_\theta^\beta(\mathcal{E}_f) < \pi_\theta^\beta(\mathcal{E}_d)\), to achieve robustness against unlearning on graph.


\begin{proposition}
\label{proposition:3.2}
For edge unlearning, edges with high influence exhibit a low predicted probability \(\pi_{\theta}(y \mid x)\), and unlearning these edges would lead to slower divergence speed and reduced impact on the retain set.
\end{proposition}

Proposition \ref{proposition:3.2} indicates \textbf{NPO is suitable for handling requests that contain a significant number of high-influence edges on edge unlearning task}. To validate the proposition, this work configures a deletion request to consist of edges characterized by high influence. As shown in Figure \ref{fig2}, as the predicted probability of high-influence edges on the forget set decreases, the counterpart on the retain set remains stable without decreasing.

\section{Methodology}
Although NPO is suitable for high-influence edge unlearning, random deletion requests are more common. Improving the general unlearning capability of NPO is challenging for graph unlearning. In this section, we propose INPO to improve the robustness of the model utility to graph unlearning.

\subsection{Fast Estimation of Edge Influence}
\label{influence fast est}
In this work, we use a remove-based approach\cite{Rong2023EfficientGNN,Stepin2021Survey,Yuan2020XGNN,Yuan2021ExplainabilitySubgraphExplorations} to estimate the influence of the nodes. Subsequently, this assessment enables us to determine the influence exerted by the edges within the graph. To express the influence of the node \(v_r \in \mathcal{V}\), we define it as:
\begin{equation}
F_{g_{\theta}}(v_r) = \sum_{\substack{i=1, i \neq r}}^{N} \left\| g_{\theta}(G)_i - g_{\theta}(G_{-v_r})_i \right\|_1 ,
\end{equation}
where \(g_{\theta}(G)_i \in \mathbb{R}^c\) (c is the number of classes)  denote the predicted class probability of node \(v_i\), and is trained on graph \(G\). \(G_{-v_r}\) is the graph that node \(v_r\) is removed.

To obtain the influence of all nodes, a direct and simple method is to alternately remove every node and predict with the trained GNN on the modified graph. The difference is the influence of all nodes. However, this brute-force way is time-consuming. Considering efficiency, we adopt the gradient information to approximate the removal-based node influence as NORA\cite{li2024fast}. NORA derives the node influence as follows:
\begin{equation}
\label{nora}
\begin{aligned}
&F_{g_{\theta}}(v_r) = \sum_{\substack{i=0}}^{L-1} (\hat{d}_r^{(L-1-i)}\hat{h}_r^{(i)}) + k_3 \cdot \delta{Topo_r} , \\
\hat{d}_r = 1 - &\frac{d_r}{(N-1)(d+\gamma)}, \, \hat{h}_r^{(i)}=\frac{d_r}{d_r+\gamma} \parallel (f_r \frac{\partial{f_r}}{\partial{h_r^{(i)}}}) \circ h_r^{(i)} \parallel_1 , \\
&\delta{Topo_r} = \sum_{i \in N(r)} \sum_{j \in N(i)}[k_1(\frac{1}{\sqrt{d_i - 1}} -\frac{1}{\sqrt{d_i}}) \\
+(1-&k_1)(\frac{1}{d_i - 1} -\frac{1}{d_i})][k_2 \frac{1}{\sqrt{d_j}} + k_2^{'}\frac{1}{d_j} + (1 - k_2 - k_2^{'})] ,
\end{aligned}
\end{equation}
where \(d_r\) and \(d\) represent the degree of the removed node \(v_r\)  and the average degree of the entire graph, respectively. \(f_r = \sum_{i=1,i \neq r}^{N} h_i^{(L)}\) is the sum of the predicted probability of all nodes except node \(v_r\) at \textit{L}-th GNN layer, and \(k_1,k_2,k_2^{'},k_3\) and \(\gamma\) are hyperparameters. \(\circ\) denotes element-wise production.

According to equation \ref{nora}, we can get the influence of all nodes for the entire graph only by a single inference. The edge \((v_i, v_j)\) influence can be expressed as:
\begin{equation}
\xi_{ij} = F_{g_{\theta}}(v_i) + F_{g_{\theta}}(v_j).
\end{equation}

Compared to the brute-force method, this gradient approximation can make a fast estimation without excessive training and computational overhead.

\subsection{Influence-Enhanced MPNN}
To improving the general unlearning capability of NPO, i.e., the deletion request consists of a randomly selected subset of edges, we redesign the massage passing mechanism in GNN to adapt to edge unlearning task. The traditional massage passing mechanism contains three components\cite{Gilmer2017NeuralMessagePassing}: the message function, the aggregate function, and the update function. The proposition \ref{proposition:3.2} shows that NPO is suitable for unlearning edges with high influence. A direct design is overwriting the massage function and enhancing the influence of low-influence edges on the forget set.

The common message function is:
\begin{equation}
m_{vu}^{(l)} = \rho_{vu}h_u^{(l-1)} ,
\end{equation}
where \(m_{vu}^{(l)}\) is the message at GNN layer \(l\), and \(\rho_{vu}=\frac{1}{\sqrt{d_v \cdot d_u}}\) is normalized weight between two nodes.

Before fine-tuning the pre-trained GNN for edge unlearning, we first use the NORA algorithm in Section \ref{influence fast est} to estimate the influence of all edges. Therefore, overwriting the massage function of unlearned model would not affect the estimation of edge influence in the graph.

After obtaining the influence of each edge through NORA, we rewrite the message function as follows:
\begin{equation}
\begin{aligned}
m_{vu}^{(l)} &= e^{q\xi_{vu}}\rho_{vu}h_u^{(l-1)}, \\
&(v, u) \in \mathcal{E}_f,
\end{aligned}
\end{equation}
 where q is a hpyerparameter. Compared to influence of edges on the retain set, the new massage function enhances the influence on the forget set. In the actual implementation, the size of unlearned edges is small, and we adopt another method to reduce the influence of edges in the retain set. The the message function is:
 \begin{equation}
\begin{aligned}
m_{vu}^{(l)} &= e^{-q\xi_{vu}}\rho_{vu}h_u^{(l-1)}, \\
&(v, u) \in \mathcal{E}_r .
\end{aligned}
\end{equation}

The new massage considers the influence of edges in the original graph and reduces impact on the retain set by amplifying the influence of unlearned edges, which makes the NPO suitable for edge unlearning. Actually, this approach mitigates the tight topological coupling by enlarging the probability difference between the forget set and the retain set.

\subsection{Topological Entropy}
\label{sec:loss}
This work focuses on parameter optimization, i.e., \textbf{unlearning fine-tuning}\cite{Hu2022LoRA}, and modifies the pre-trained model parameters by different objectives. Based on the objective of unlearned graph model, we categorize existing methods\cite{Maini2024TOFU,Shi2025MUSE} into two paradigms: \textbf{preserve model utility} and \textbf{improve forget quality}.

For preserving utility of model, We consider the following two methods:
\begin{itemize}
\item \textbf{Gradient Descent (GD)} simply uses the training CE loss to perform gradient descent on the retain set, as follows:
\begin{equation}
L_{GD}(\mathcal{E}_r; \theta) = \mathbb{E}_{(x, y) \sim \mathcal{E}_r}[-log \, \pi_\theta(y \mid x)].
\end{equation}
\item \textbf{Kullback-Leibler Divergence (KL)}\cite{Haldar2025LLM} is to minimize the difference of the prediction distribution of the unlearned model and the reference model on the retain set, as follows:
\begin{equation}
\quad \quad \quad  L_{KL}(\mathcal{E}_r; \theta) = \mathbb{E}_{(x, y) \sim \mathcal{E}_r}[ KL(\pi_\theta(y \mid x) \, || \, \pi_{ref}(y \mid x))].
\end{equation}
\end{itemize}

For improving forget quality, We consider the following two methods:
\begin{itemize}
\item \textbf{Gradient Ascent (GA)} maximize the CE loss loss on the forget set, as follows:
\begin{equation}
L_{GA}(\mathcal{E}_f; \theta) = -\mathbb{E}_{(x, y) \sim \mathcal{E}_f}[-log \, \pi_\theta(y \mid x)].
\end{equation}
\item \textbf{Direct Preference Optimization (DPO)} use prediction on the forget set as negative samples and random prediction on the retain set as positive samples.
\end{itemize}

To incorporate the properties of the graph into the edge unlearning process, we propose a new optimization objective from a topological perspective. Inspired by the Neighborhood Influence property which the embedding of the neighboring subgraph remains largely unchanged before and after the edge deletion in GNNDelete. We directly average embedding \(h_i\) and \(h_j\) at layer \(L\) to obtain the distribution of the local subgraph around that edge \(e_{ij}\), it is:
\begin{equation}
G_{ij} = \frac{1}{2}(h_i^{(L)} + h_j^{(L)}) ,
\end{equation}
where \(G_{ij}\) represents the embedding of the \textit{L}-hop local structure.

To ensure that edge unlearning does not cause significant changes to the neighboring nodes, we propose topological entropy as an optimization objective as following:
\begin{equation}
TE_{ij} = -\sum G_{ij}^{ref} \, log(G_{ij}) , \quad e_{ij} \in \mathcal{E}_f , 
\end{equation}
where \(G_{ij}^{ref}\) represents the pre-trained embedding of the L-hop local structure.

Finally, We employ a holistic loss function to optimize two losses:
\begin{equation}
\label{overall-loss}
Loss = \lambda_1 L_{NPO} + \lambda_2 GD + \lambda_3 TE, 
\end{equation}
where \(\lambda_1, \lambda_2, \lambda_3\) are weights associated with forget quality and model utility. These weights decide whether the model is more inclined to improve forget quality or preserve model utility.

\subsection{Complexity Analysis}
Compared to other unlearning fine-tuning mdoels, our model adds a single inference to calculate the influence of edges, and this overhead is negligible. We list the time and space complexities\cite{Wu2020GNN,Blakely2019Time} in Table \ref{tab:complexity}. \(E\) denotes the number of edges, and \(F\) denotes the feature dimension.  It is easy to see that the order of complexity remains unchanged, the time complexity and space complexity are \(O(LEF + LNF^2)\) and \(O(E + LF^2 + LNF)\), respectively. 

\begin{table}
  \caption{Complexity comparison}
  \label{tab:complexity}
  \begin{tabular}{ccc}
    \toprule
    Method&Fine-tuning&Ours\\
    \midrule
    Time & \(LEF + LNF^2\) & \(2LEF + LNF^2\)\\
    Space & \(E + LF^2 + LNF\) & \(2E + LF^2 + LNF\)\\
  \bottomrule
\end{tabular}
\end{table}

\begin{table*}[t!]
\vspace{-10pt}
\caption{Comparison results of our model with self-designed fine-tuning methods. In each column, the best result is indicated in red, while the runner-up result is marked with blue. The pre-traing based on link prediction task.}
\label{fig:overall-performance-self}
\vspace{-5pt}
\begin{center}
\begin{small}
\setlength{\tabcolsep}{12pt} 
\begin{tabular}{lcccccccccc}
\toprule
       & \multicolumn{5}{c}{DBLP} & \multicolumn{5}{c}{Cora} \\
\cmidrule(lr){2-6} \cmidrule(lr){7-11} 
Model  & \multicolumn{2}{c}{$\mathcal{E}_r$} & \multicolumn{3}{c}{$\mathcal{E}_f$} & \multicolumn{2}{c}{$\mathcal{E}_r$} & \multicolumn{3}{c}{$\mathcal{E}_f$} \\
\cmidrule(lr){2-3} \cmidrule(lr){4-6} \cmidrule(lr){7-8} \cmidrule(lr){9-11}
       & AUC & AP & AUC & AP & MI Ratio & AUC & AP & AUC & AP & MI Ratio \\
\midrule
GA          & 0.6787       & 0.6156       & 0.6046       & 0.5589       & 1.30       & 0.5106       & 0.5799       & 0.6025     & 0.5766    & 2.77      \\
GA+GD       & 0.6122       & 0.6038       & 0.8498       & \textcolor{blue}{0.8445}       & 2.13       & 0.5181       & 0.5851       & 0.6153     & 0.5923    & 2.74       \\
GA+KL       & 0.6788       & 0.6157       & 0.6046       & 0.5589       & 1.30       & 0.6668         & 0.6125       & 0.5487   & 0.5266    & 1.21       \\
DPO         & 0.6501       & 0.6718       & \textcolor{blue}{0.8639}       & 0.8312       & 2.60       & 0.5495       & 0.6008     & 0.6718   & 0.6543    & 2.76       \\
DPO+GD      & 0.9432       & 0.9352       & 0.4842       & 0.4864       & 1.02       & 0.7256       & 0.7141     & 0.6311   & 0.5706    & 1.54       \\
DPO+KL      & 0.8245       & 0.7713       & 0.4657       & 0.4819       & 1.00       & 0.7104       & 0.6434     & 0.5046   & 0.5023    & 1.00       \\
\midrule
NPO         & 0.9002       & 0.9027       & 0.7913       & 0.8038       & 1.79       & 0.8996       & 0.9015     & \textcolor{blue}{0.7142}   & \textcolor{blue}{0.7036}    & 1.84       \\
\textbf{INPO}         & 0.8853       & 0.8852       & \textcolor{red}{0.9037}       & \textcolor{red}{0.9010}       & \colorbox{yellow}{\textbf{1.59}}       & 0.8973       & 0.8916       & \textcolor{red}{0.9058}   & \textcolor{red}{0.8885}    & \colorbox{yellow}{\textbf{1.61}}       \\
\bottomrule
\end{tabular}
\end{small}
\end{center}
\end{table*}

\begin{table*}[t!]
\caption{Comparison results of our model with advanced fine-tuning methods. In each column, the best result is indicated in red, while the runner-up result is marked with blue, and the third palce is marked with orange. Evaluation: link prediction.}
\label{fig:overall-performance-advanced}
\begin{center}
\begin{small}
\setlength{\tabcolsep}{12pt} 
\begin{tabular}{lcccccccccc}
\toprule
       & \multicolumn{5}{c}{DBLP} & \multicolumn{5}{c}{Cora} \\
\cmidrule(lr){2-6} \cmidrule(lr){7-11} 
Model  & \multicolumn{2}{c}{$\mathcal{E}_r$} & \multicolumn{3}{c}{$\mathcal{E}_f$} & \multicolumn{2}{c}{$\mathcal{E}_r$} & \multicolumn{3}{c}{$\mathcal{E}_f$} \\
\cmidrule(lr){2-3} \cmidrule(lr){4-6} \cmidrule(lr){7-8} \cmidrule(lr){9-11}
       & AUC & AP & AUC & AP & MI Ratio & AUC & AP & AUC & AP & MI Ratio \\
\midrule
Retrain     & 0.9614       & 0.9645       & 0.5153       & 0.5131       & 1.05       & 0.9364       & 0.9355       & 0.4818     & 0.4867    & 1.09      \\
GA          & 0.6787       & 0.6156       & 0.6046       & 0.5589       & 1.30       & 0.5106       & 0.5799       & 0.6025     & 0.5766    & 2.77      \\
GIF         & 0.9688       & 0.9714       & 0.5217       & 0.5168       & 1.03       & 0.9678       & 0.9668       & 0.4913     & 0.4937    & 1.03       \\
GNNDelete   & 0.9573       & 0.9601       & \textcolor{blue}{0.9731}       & \textcolor{blue}{0.9754}       & \colorbox{yellow}{1.69}       & 0.9609       & 0.9609       & \textcolor{blue}{0.9797}     & \textcolor{blue}{0.9834}    & \colorbox{yellow}{1.75}       \\
UtU         & 0.9687       & 0.9714       & 0.5158       & 0.5098       & 1.03       & 0.9677       & 0.9668       & 0.4965     & 0.4924    & 1.03        \\
\midrule
NPO         & 0.9002       & 0.9027       & 0.7913       & 0.8038       & 1.79       & 0.8996       & 0.9015     & 0.7142   & 0.7036    & 1.84       \\
\textbf{INPO}         & 0.8853       & 0.8852       & \textcolor{orange}{0.9037}       & \textcolor{orange}{0.9010}       & \colorbox{yellow}{1.59}      & 0.8973       & 0.8916       & \textcolor{orange}{0.9058}   & \textcolor{orange}{0.8885}    & \colorbox{yellow}{1.61}       \\
\textbf{INPO-S}         & 0.9533       & 0.9554       & \textcolor{red}{0.9809}       & \textcolor{red}{0.9823}       & \colorbox{yellow}{\textbf{1.80}}       & \textcolor{red}{0.9613}       & \textcolor{red}{0.9613}       & \textcolor{red}{0.9802}   & \textcolor{red}{0.9836}    & \colorbox{yellow}{\textbf{1.79}}  \\
\bottomrule
\end{tabular}
\end{small}
\end{center}
\end{table*}

\section{Experiments}
\subsection{Experimental Settings}
\subsubsection{Datasets.}   To thoroughly validate the effectiveness of our model and ensure a comprehensive generalization evaluation, we used five real-world datasets\cite{Aleksandar2018deep, hu2024open}: Cora, PubMed, DBLP, CS, OGB-Collab. 


\subsubsection{Baseline Models.} In our experiments, we select 4 advanced and 5 self-designed methods based on the loss combination discussed in Section \ref{sec:loss} as baselines for performance comparison. The description of these baselines is as follows.

\textbf{Advanced Fine-tuning Methods for Edge Unlearning.}
\begin{itemize}
\item \textbf{Retrain}\cite{Liu2022TheRight}. This method, while straightforward, is inefficient as it requires retraining models from scratch to unlearn specific edges. 
\item \textbf{GIF}\cite{Wu2023GIF}. This method accurately estimates parameter changes by designing influence functions to directly modify the parameters for edge unlearning.
\item \textbf{GNNDelete}\cite{Cheng2023}. This method achieves unlearning by approximating the representation of edges to be forgotten to those that did not exist in the pretrained model, while keeping the neighbors' representations minimally changed. 
\item \textbf{UtU}\cite{Tan2024Unlink}. Compared to GNNDelete, it only uses the graph after edge deletion for a single inference.
\end{itemize}


\textbf{Self-designed Fine-tuning Methods for Edge Unlearning.}
\begin{itemize}
\item \textbf{GA+GD}\cite{Maini2024TOFU}. This method use GA loss on the forget set and GD loss on the retain set as optimization objective.
\item \textbf{GA+KL}\cite{Maini2024TOFU}. This method use GA loss on the forget set and KL loss on the retain set as optimization objective.
\item \textbf{DPO}\cite{Rafailov2023}. This method treat the edge unlearning as a preference optimization problem. We use predicted probability on the forget set as negative
samples and random probability on the retain set as positive samples to perform preference optimization.
\item \textbf{DPO+GD}\cite{Yuan2025}. This method use DPO loss and GD loss on the retain set as optimization objective.
\item \textbf{DPO+KL}\cite{Yuan2025}. This method use DPO loss and KL loss on the retain set as optimization objective.
\end{itemize}

\subsubsection{Evaluation Metrics.} To measure the effectiveness of our model, we use model utility and forget quality as metrics. The model utility refers to its ability to maintain the original inference capability after unlearning, measured by AUC and AP on the retain set. The forget quality measured by AUC and AP on the forget set. \(\text{AP} = \sum_n (R_n - R_{n-1}) \cdot P_n\) where \(P_n\) and \(R_n\) are the precision and recall at the \textit{n}-th threshold. AUC is the area under the Receiver Operating Characteristic Curve. To evaluate whether the model has truly achieved forgetting, we use \textbf{the probability of edge(\(e \in \mathcal{E}_f \)) existence} after unlearning as another metric of forgetting quality, i.e., \(p_f(avg)\). MI Ratio\cite{Olatunji2021MembershipInferenceGNN} is a commonly used metric for measuring model forget quality, which quantifies the success rate of a Membership Inference (MI) attack\cite{Yeom2019PrivacyRisk,Sablayrolles2019WhiteboxBlackbox}, by calculating the ratio of presence probability of \(\mathcal{E}_f\) before and after the deletion operator.

\subsubsection{Implementation Details.} We evaluate the effectiveness of our model on edge unlearning tasks. We perform experiments on two settings: (1) the deletion request consists of edges with high influence and (2) the deletion request consists of random edges. To perform edge unlearning tasks, The proportion of edges we delete is \(0.5\%\). We conducted all experiments 5 times and reported average value, ignored the variance because they were extremely small.

\subsection{Overall Performance Experiments}
\textbf{Analysis on the baselines}. In Table \ref{fig:overall-performance-self}, \ref{fig:overall-performance-advanced} and \ref{fig:overall-performance-prob}, we summarize the overall performance of INPO and the baselines. We observe that the forget quality of DPO and NPO constantly surpasses most advanced fine-tuning methods, showing the effect of preference optimization for graph edge unlearning. Also, we found that GD loss can improve the performance on the retain set. Moreover, most baselines is hard to strike a balance between model utility and the quality of forgetting, except for GNNDelete. 
An interesting observation is that the MI ratio and \(\frac{pr}{pf}\) of GNNDelete are relatively low, indicating that it does not truly unlearn the edges on the forget set. The higher edge prediction probability on the forget set also indicates this. In conclusion, the limitations of baselines hinder their ability to achieve consistent success.

\begin{table}[h]
\caption{The average predicted probability of our model with baseline methods on the retain set and the forget set. Table 4 has all hyper-parameters consistent with those in Table 3. In each column, the best result is indicated in red, while the runner-up result is marked with blue.}
\label{fig:overall-performance-prob}
\begin{center}
\begin{small}
\setlength{\tabcolsep}{6pt} 
\begin{tabular}{lcccccc}
\toprule
       & \multicolumn{3}{c}{DBLP} & \multicolumn{3}{c}{Cora} \\
\cmidrule(lr){2-4} \cmidrule(lr){5-7} 
Model  & \multicolumn{1}{c}{$p_f$} & \multicolumn{1}{c}{$p_r$} & \multicolumn{1}{c}{$\frac{pr}{p_f}$} & \multicolumn{1}{c}{$p_f$} & \multicolumn{1}{c}{$p_r$} & \multicolumn{1}{c}{$\frac{pr}{p_f}$} \\
\midrule
Retrain     & 0.9305       & 0.9010     & 0.97       & 0.8971       & 0.8626    & 0.96             \\
GIF         & 0.9504       & 0.9129     & 0.96       & 0.9492       & 0.9093    & 0.96              \\
GNNDelete   & \colorbox{yellow}{0.5814}       & \colorbox{pink}{0.8517}     & \textcolor{blue}{1.46}       & \colorbox{yellow}{0.5595}       & \colorbox{pink}{0.8524}    & \textcolor{blue}{1.52}              \\
UtU         & 0.9496       & 0.9122     & 0.96       & 0.9472       & 0.9076    & 0.96               \\
NPO         & 0.5475       & 0.5910     & 1.08       & 0.5323       & 0.5476    & 1.03             \\
\textbf{INPO}         & \colorbox{yellow}{0.6154}       & \colorbox{pink}{0.8143}       & 1.32       & \colorbox{yellow}{0.6070}       & \colorbox{pink}{0.8036}   & 1.32       \\
\textbf{INPO-S}         & \colorbox{yellow}{0.5451}       & \colorbox{pink}{0.8388}       & \textcolor{red}{1.54}       & \colorbox{yellow}{0.5473}       & \colorbox{pink}{0.8562}   & \textcolor{red}{1.56}        \\
\bottomrule
\end{tabular}
\end{small}
\end{center}
\end{table}

\begin{table*}[h]
\caption{Ablation results of our model.}
\label{fig:ablation-performance}
\begin{center}
\begin{small}
\setlength{\tabcolsep}{10pt} 
\begin{tabular}{lcccccccccc}
\toprule
       & \multicolumn{5}{c}{DBLP} & \multicolumn{5}{c}{Cora} \\
\cmidrule(lr){2-6} \cmidrule(lr){7-11} 
Model  & \multicolumn{2}{c}{$\mathcal{E}_r$} & \multicolumn{3}{c}{$\mathcal{E}_f$} & \multicolumn{2}{c}{$\mathcal{E}_r$} & \multicolumn{3}{c}{$\mathcal{E}_f$} \\
\cmidrule(lr){2-3} \cmidrule(lr){4-6} \cmidrule(lr){7-8} \cmidrule(lr){9-11}
       & AUC & AP & AUC & AP & MI Ratio & AUC & AP & AUC & AP & MI Ratio \\
\midrule
NPO+GD      & 0.8106       & 0.8234       & 0.8549       & 0.8167       & 1.91       & 0.7923       & 0.7809     & 0.9057   & 0.8934    & 1.73    \\
NPO+IMPNN   & 0.8708       & 0.8693       & 0.8150       & 0.8248       & 1.92       & 0.8375       & 0.8449     & 0.7822   & 0.7845    & 1.93      \\
NPO+TE      & 0.8977       & 0.8966       & 0.8323       & 0.8501       & 1.86       & 0.8734       & 0.8767     & 0.7540   & 0.7561    & 1.90    \\
NPO+GD+IMPNN   & 0.8231    & 0.8253       & 0.8792       & 0.8789       & 1.86       & 0.8543       & 0.8456       & 0.8847     & 0.8716     & 1.61       \\
NPO+GD+TE   & 0.8384       & 0.8362       & 0.9220       & 0.9227       & 1.76       & 0.8562       & 0.8465       & 0.9103     & 0.8954     & 1.72       \\
\textbf{INPO}         & 0.8853       & 0.8852       & 0.9037       & 0.9010       & 1.59    & 0.8973       & 0.8916       & 0.9058   & 0.8885    & 1.61       \\
\bottomrule
\end{tabular}
\end{small}
\end{center}
\end{table*}

\begin{figure}[t]
  \centering
  \includegraphics[width=0.8\linewidth]{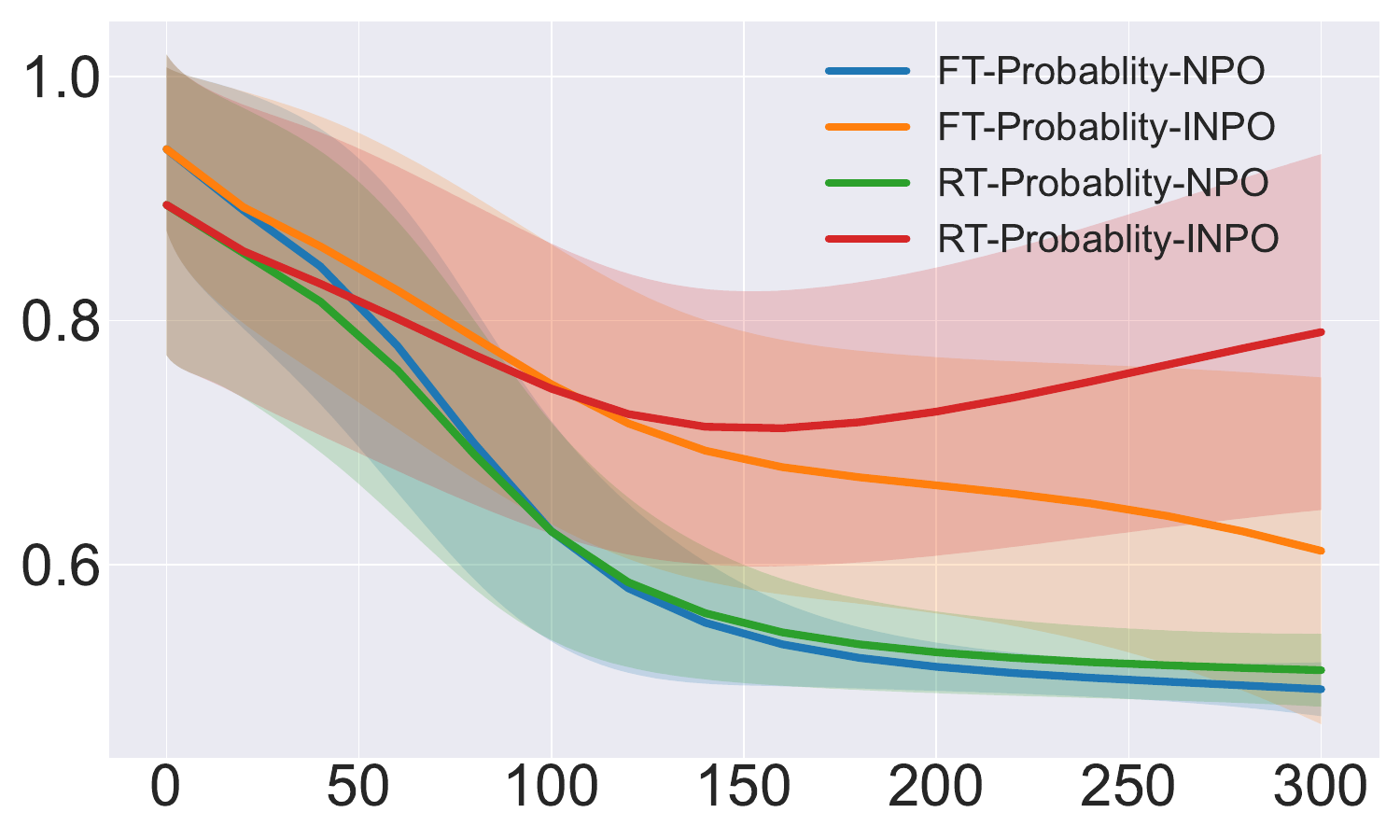}
  \caption{The prediction probability change curve of NPO and INPO on Cora validation dataset.}
  \label{fig:ablation-p}
\end{figure}

\begin{figure}[t]
  \centering
  \includegraphics[width=0.8\linewidth]{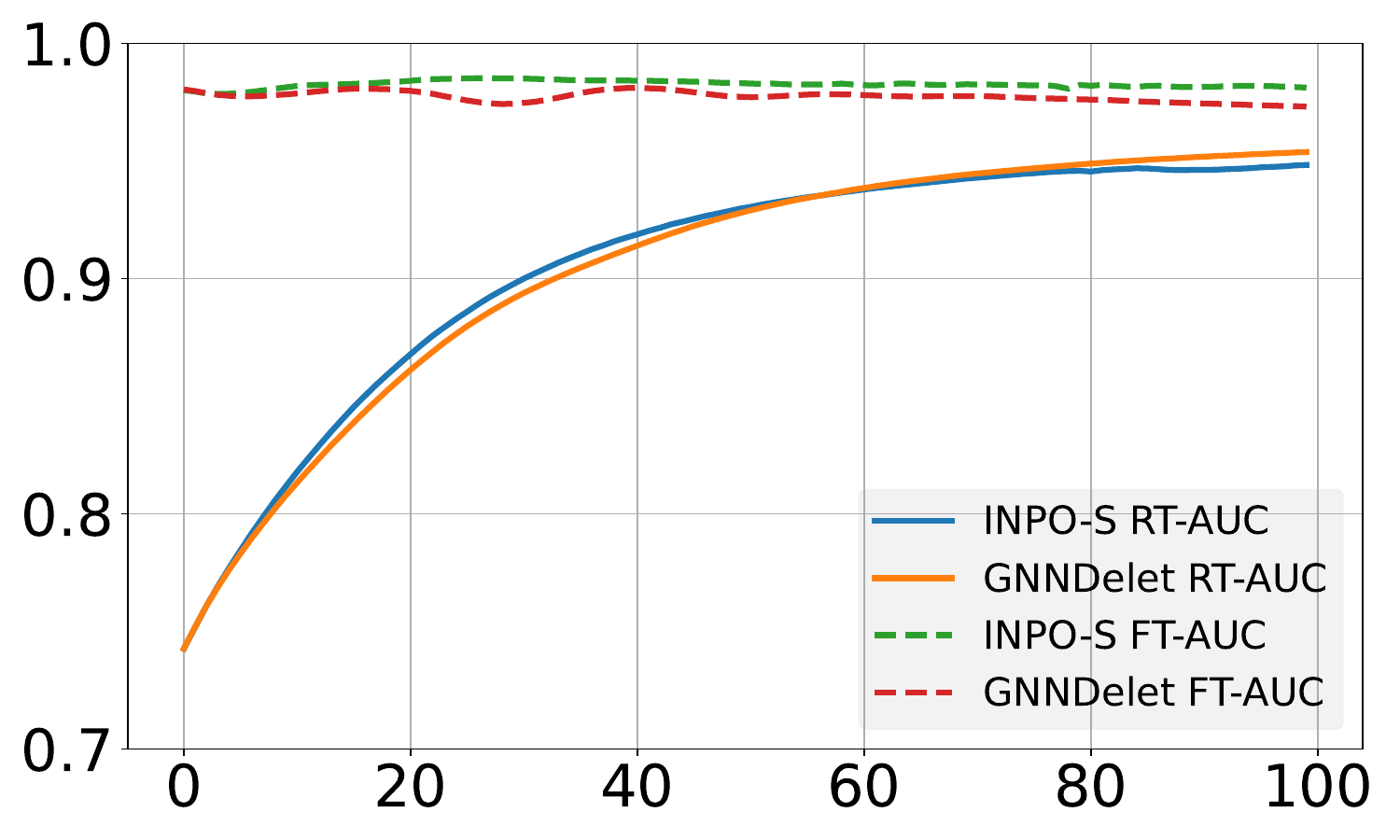}
  \caption{The AUC change curve of INPO-S and GNNDelete on retain set and forget set for DBLP validation dataset.}
  \label{fig:gnndelete-inpo}
  \vspace{-10px}
\end{figure}

\textbf{The effectiveness of INPO}. Overall, INPO outperforms most of the baselines in terms of model utility and forget quality, and INPO-S achieves state-of-the-art performance on all forget quality metrics while maintaining the model’s utility. In particular, we obtain large forget quality gains over the best baseline in two datasets by \(6.5\%\) and \(2.3\%\)  for MI Ratio, respectively. Additionally, compared to GNNDelete, which was previously the baseline with the best balance between model utility and the quality of forgetting, INPO-S improves \(\frac{p_r}{p_f}\) by \(5.5\%\) and \(2.6\%\) on two datasets. Notably, INPO-S achieves a perfect model utility that is essentially the same as GNNDelete. This evidence suggests that INPO is able to achieve SOTA edge unlearning, and the decreased prediction probability for edges on the forget set provide a specific explanation. More results of experiments are given in Supplementary Materials A.2.

\textbf{Comparison between NPO and INPO}. As shown in Figure \ref{fig:ablation-p}, we found that INPO effectively mitigates the impact of the unlearning process on model utility, which is missing in NPO. From the gradient perspective, IMPNN reduces the adaptive coefficient \(S_{\theta}(x,y)\), thereby minimizing impact on model utility, while maintaining model utility through TE loss. \textbf{The substantial improved \(\frac{p_r}{p_f}\) indicates effectiveness of mitigating the tight topological coupling}. Further experimental results are provided in the ablation study and Supplementary Materials A.2(Figure 3). 

\textbf{Comparison between GNNDelete and INPO-S}. INPO-S refers to a method that incorporates additional parameters \textbf{initialized to zero} for forgetting, similar to GNNDelete. The key distinction is that INPO-S does not utilize the \textbf{Deleted Edge Consistency} loss employed by GNNDelete for the forgetting process. As shown in Figure \ref{fig:gnndelete-inpo}, we found that INPO performs better and more stably in maintaining the forgetting capability.

\begin{figure}[t]
  \centering
  \includegraphics[width=0.8\linewidth]{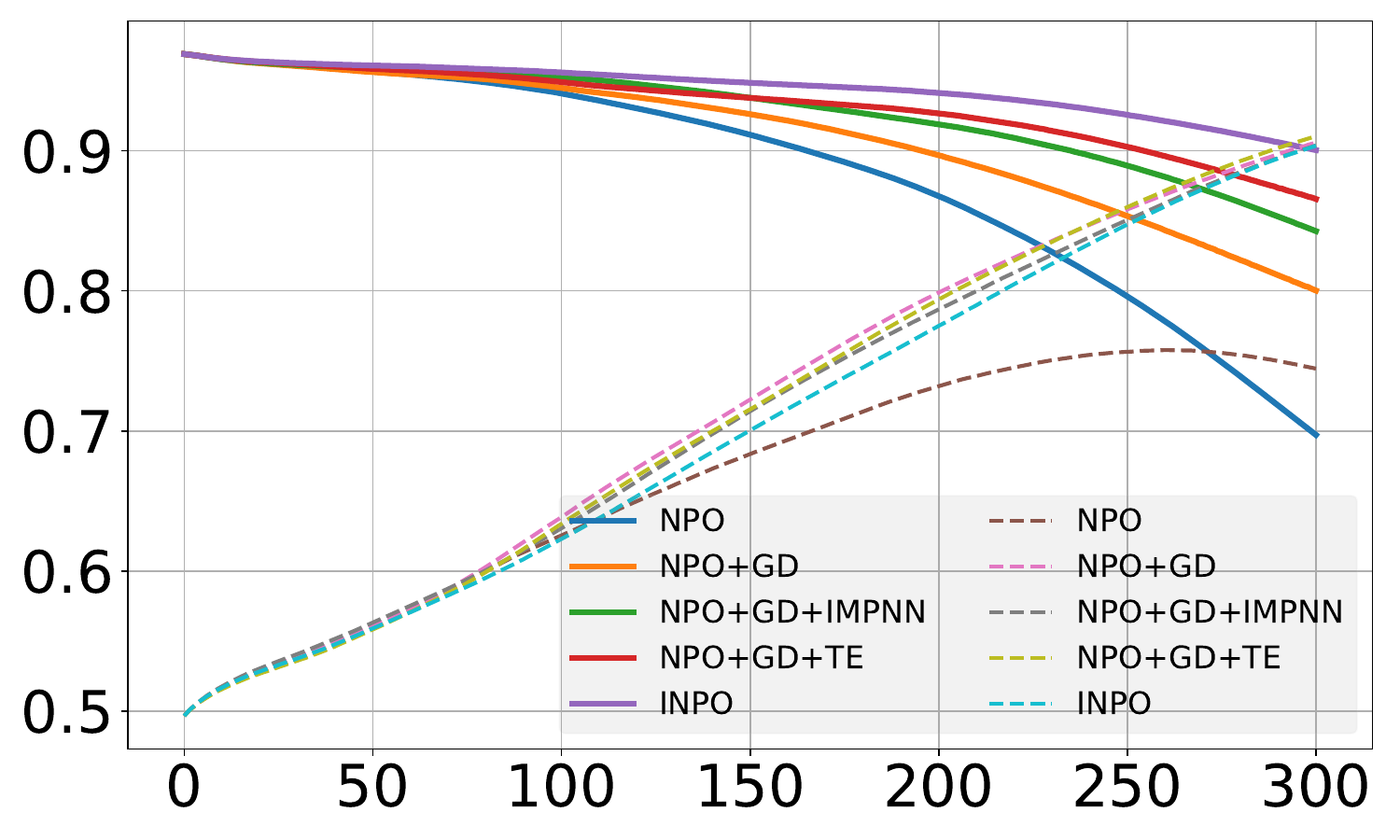}
  \caption{The AUC change curve of all ablation models on Cora validation dataset. The solid lines denotes AUC on retain set and the dashed lines represents AUC on forget set.}
  \label{fig:ablation}
  \vspace{-10px}
\end{figure}

\subsection{Ablation Experiment}
Here we empirically dissect the contribution of (1) GD loss, (2) redesigned MPNN, and (3) topological entropy regularization. We proposed five ablations models respectively:
\begin{itemize}
\item \textbf{NPO-GD}, which uses GD loss as a regularization term for NPO loss. 
\item \textbf{NPO-IMPNN}, which replaces the message passing mechanism with influence-based message function.
\item \textbf{NPO-TE}, which uses topological entropy as a regularization term for NPO loss.
\item \textbf{NPO-GD-IMPNN}, which uses GD loss regularization and influence-based message function.
\item \textbf{NPO-GD-TE}, which uses GD loss and topological entropy regularization.
\end{itemize}
\textbf{Ablation results}.  In Table \ref{fig:ablation-performance}, we report the ablation results. By comparing NPO+GD and NPO+GD+IMPNN, we discover that can effectively maintain AUC and AP on the retain set while enhancing the forget quality on the forget set for fine-tune. This result demonstrates the effectiveness of redesigned influence-based message function. Further, the comparison between NPO+GD and NPO+GD+TE implies that, as the unlearn process progresses, it can still effectively maintain performance on the retain set. This result demonstrates the effectiveness of topological entropy regularization. Overall, these three ablation models justify the efficacy of our framework.


\textbf{Trade-off between utility and forget quality}. Here we are interested in the changes of utility and forget quality during the optimization process, and we visualize the trade-off process on Cora dataset. As shown in Figure \ref{fig:ablation}, we observe that: (1) In terms of model utility and the forget quality, the three ablation models and INPO significantly outperform the original NPO. (2) INPO is largely consistent with the three ablation models(except for NPO) for the forget quality, but it excels in maintaining AUC without large decline.

\subsection{Robustness Analysis Experiments}
In this section, we delve into the robustness of our framework from three perspectives: \(\beta\) in Equation \ref{eq:npo}, \(\lambda_1\) in Equation \ref{overall-loss} and \(\lambda_3\) in Equation \ref{overall-loss}. The analysis was conducted using the Cora dataset.

\begin{figure}[t]
  \centering
  \includegraphics[width=0.8\linewidth]{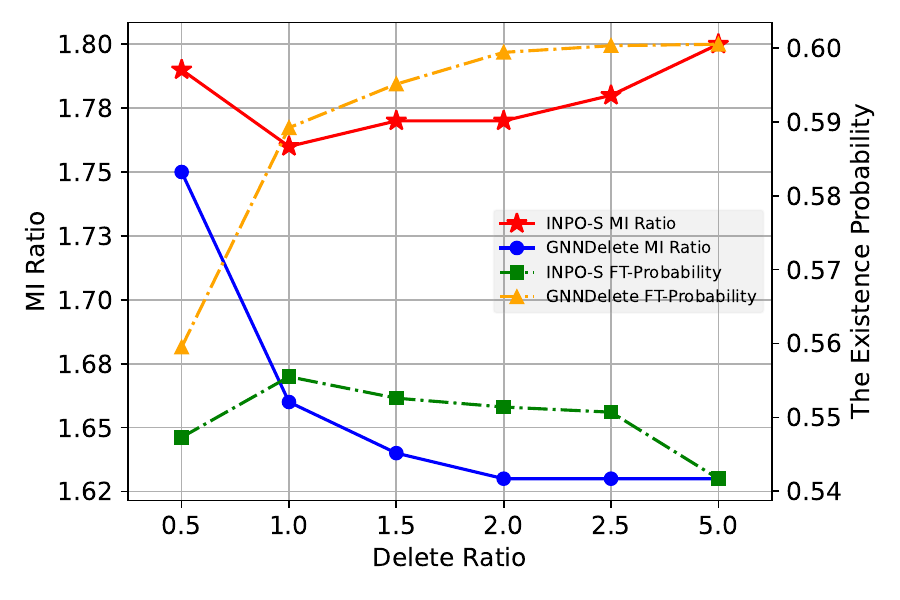}
  \caption{MI Patio performance and the probability of edge \((e\in \mathcal{E}_f)\) existence for different delete ratio(\(\%\)), and a lower probability of edge existence indicates better unlearning.}
  \label{fig:ratio}
\end{figure}

\textbf{The robustness to different delete ratio}. As shown in Figure \ref{fig:ratio}, INPO-S significantly outperforms the current best baseline in both MI Ratio and the probability of edge existence. It is noteworthy that our MI Ratio not only outperforms the baseline, but also \textbf{does not show a decline as the delete ratio increases}, unlike GNNDelete. In fact, our model demonstrates a slight improvement with higher delete ratio. Additionally, \textbf{the probability of the edges we aim to forget does not increase as the delete ratio grows}, indicating that INPO-S achieves true forgetting even at higher delete ratio. These findings indicate that our model exhibits strong robustness across different deletion ratios.

\begin{figure}[t]
  \centering
  \includegraphics[width=0.8\linewidth]{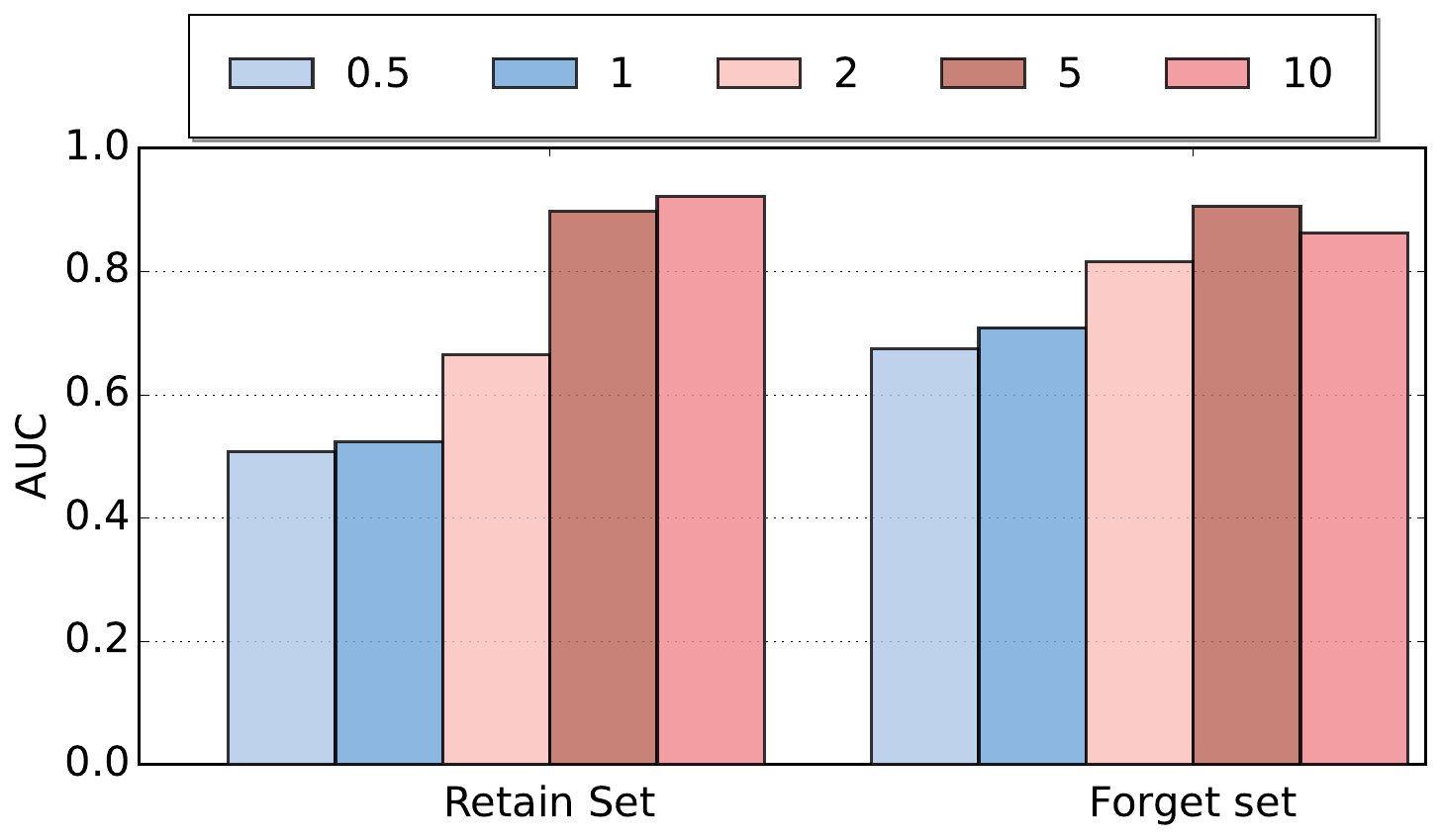}
  \caption{AUC Performance on the retain set and the forget set at different \(\beta\).}
  \label{fig:sens:beta}
\end{figure}

\begin{figure}[t]
    \centering
    \begin{subfigure}[b]{0.23\textwidth}
        \includegraphics[width=\textwidth]{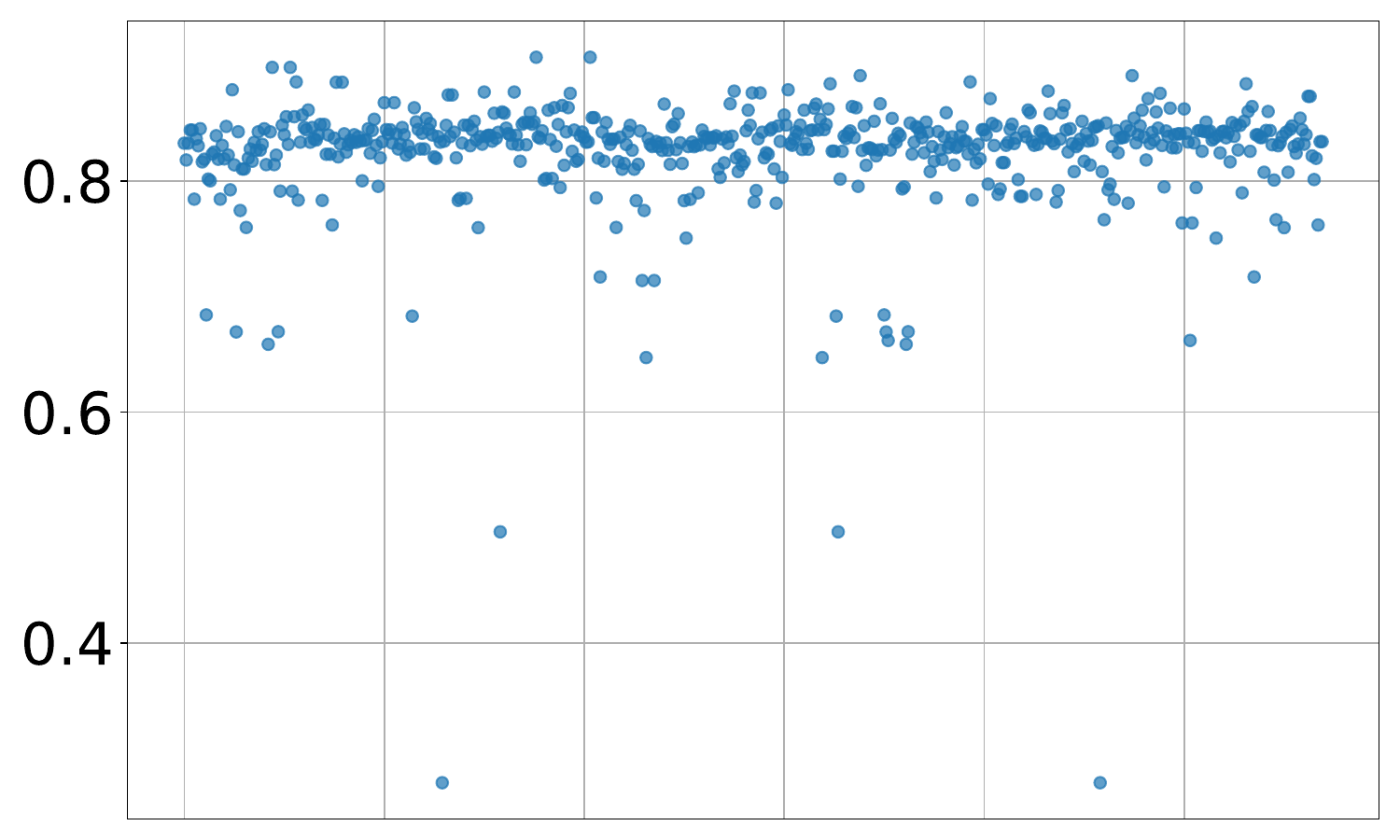}
        \caption{\(S_\theta(x, y)\) for \(\beta=0.5\).}
        \label{beta:divergence:left}
    \end{subfigure}
    \hfill
    \begin{subfigure}[b]{0.23\textwidth}
        \includegraphics[width=\textwidth]{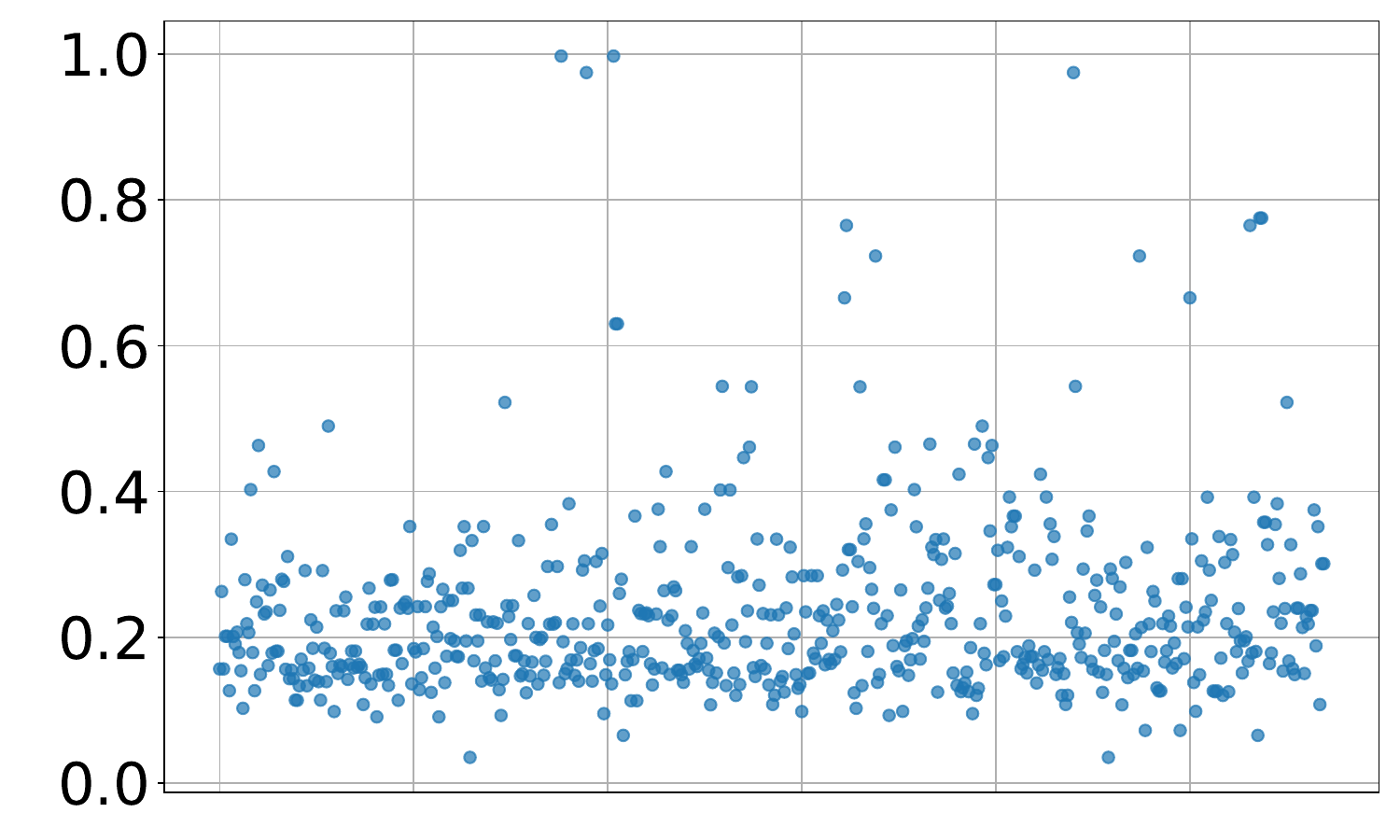}
        \caption{\(S_\theta(x, y)\) for \(\beta=5\). }
        \label{beta:divergence:right}
    \end{subfigure}
    \caption{The coefficient \(S_\theta(x, y)\) at epoch 200 for \(\beta=0.5\) and \(\beta=5\). A data point represents an edge to be forgotten.}
    \label{beta:divergence:both}
\end{figure}

\textbf{The impact of \(\beta\)}.
In Figure \ref{fig:sens:beta}, We show the impact of the hyperparameter \(\beta\) on INPO's performance, i.e., AUC on the retain set and forget set. We observe that as the number of \(\beta\) increases, the AUC on the retain set gradually improves. However, the AUC on the forget set reaches a plateau when \(\beta=5\), and then begins to decline. This difference is caused by the divergence speed, which is related to the adaptive coefficient \(S_\theta(x, y)=2 \pi_\theta^\beta(y \mid x) / [ \pi_\theta^\beta(y \mid x) + \pi_{ref}^\beta(y \mid x) ]\). When \(\beta\) is small, the divergence rate of the entire process becomes too rapid, as \(\pi_{ref}\) is directly obtained from the pre-trained model. As shown in Figure \ref{beta:divergence:both}, the divergence speed is fast when \(\beta=0.5\), which would lead to the model utility decreasing quickly. On the other hand, an overly large \(\beta\) leads to an excessively low divergence speed, which also results in a decline in the forget quality.

\textbf{The impact of \(\lambda_1\)}. As shown in Supplementary Materials Figure 1, both too small and too large \(\lambda_1\) can lead to poor AUC on forget set. An overly small NPO loss can lead to ineffectiveness of our model, thereby preventing the model from unlearning.

\textbf{The impact of \(\lambda_3\)}. As depicted in Supplementary Materials Figure 2, owing to the incorporation of \textbf{topological entropy regularization} in INPO, we investigated the influence of TE loss on model performance. The results affirm the robustness of TE loss in our model, and values between 0.5 and 0.8 are all reasonable. Different values of \(\lambda_3\) have little difference on overall performance, with only larger values of \(\lambda_3\) causing a slight decrease in AUC on the forget set. In summary, topological entropy regularization is useful and robust.

\section{Conclusion}
To improve the robustness of the model utility to the unlearning process, we propose INPO that amplify the effects of low-influence edges on the forget set to achieve topological decoupling and topological entropy loss to avoid excessive information loss in the local structure during unlearning. Extensive experiments conducted on five real-world datasets demonstrate effectiveness of our model and achieve SOTA performance on all forget quality metrics.

\begin{acks}
This research is funded by National Natural Science Foundation of China (No. 62406347 and No. 62202302) . This research is supported by the Hong Kong RGC Theme-based Research Scheme T32-615-24/R.
\end{acks}

\bibliographystyle{ACM-Reference-Format}
\bibliography{sample-base}


\begin{thebibliography}{61}


\ifx \showCODEN    \undefined \def \showCODEN     #1{\unskip}     \fi
\ifx \showISBNx    \undefined \def \showISBNx     #1{\unskip}     \fi
\ifx \showISBNxiii \undefined \def \showISBNxiii  #1{\unskip}     \fi
\ifx \showISSN     \undefined \def \showISSN      #1{\unskip}     \fi
\ifx \showLCCN     \undefined \def \showLCCN      #1{\unskip}     \fi
\ifx \shownote     \undefined \def \shownote      #1{#1}          \fi
\ifx \showarticletitle \undefined \def \showarticletitle #1{#1}   \fi
\ifx \showURL      \undefined \def \showURL       {\relax}        \fi
\providecommand\bibfield[2]{#2}
\providecommand\bibinfo[2]{#2}
\providecommand\natexlab[1]{#1}
\providecommand\showeprint[2][]{arXiv:#2}

\bibitem[Bojchevski and Günnemann(2018)]%
        {Aleksandar2018deep}
\bibfield{author}{\bibinfo{person}{Aleksandar Bojchevski} {and} \bibinfo{person}{Stephan Günnemann}.} \bibinfo{year}{2018}\natexlab{}.
\newblock \showarticletitle{Deep gaussian embedding of graphs: Unsupervised inductive learning via ranking}. In \bibinfo{booktitle}{\emph{ICLR}}.
\newblock


\bibitem[CCPA(2018)]%
        {CCPA2018}
\bibfield{author}{\bibinfo{person}{CCPA}.} \bibinfo{year}{2018}\natexlab{}.
\newblock \bibinfo{title}{California Consumer Privacy Act of 2018}.
\newblock \bibinfo{howpublished}{\url{https://leginfo.legislature.ca.gov/faces/billTextClient.xhtml?bill_id=201720180AB375}}.
\newblock
\newblock
\shownote{AB-375, Signed into law on June 28, 2018}.


\bibitem[Chen et~al\mbox{.}(2022)]%
        {Chen2022GraphUnlearning}
\bibfield{author}{\bibinfo{person}{Min Chen}, \bibinfo{person}{Zhikun Zhang}, \bibinfo{person}{Tianhao Wang}, \bibinfo{person}{Michael Backes}, \bibinfo{person}{Mathias Humbert}, {and} \bibinfo{person}{Yang Zhang}.} \bibinfo{year}{2022}\natexlab{}.
\newblock \showarticletitle{Graph unlearning}. In \bibinfo{booktitle}{\emph{Proceedings of the ACM SIGSAC Conference on Computer and Communications Security}}.
\newblock


\bibitem[Chen et~al\mbox{.}(2024)]%
        {Chen2024link}
\bibfield{author}{\bibinfo{person}{Xiaolong Chen}, \bibinfo{person}{Yifan Song}, {and} \bibinfo{person}{Jing Tang}.} \bibinfo{year}{2024}\natexlab{}.
\newblock \showarticletitle{Link Recommendation to Augment Influence Diffusion with Provable Guarantees}. In \bibinfo{booktitle}{\emph{Proceedings of the ACM Web Conference 2024 (WWW '24)}}. \bibinfo{publisher}{Association for Computing Machinery}, \bibinfo{pages}{2509--2518}.
\newblock


\bibitem[Cheng et~al\mbox{.}(2023)]%
        {Cheng2023}
\bibfield{author}{\bibinfo{person}{Jiali Cheng}, \bibinfo{person}{George Dasoulas}, \bibinfo{person}{Huan He}, \bibinfo{person}{Chirag Agarwal}, {and} \bibinfo{person}{Marinka Zitnik}.} \bibinfo{year}{2023}\natexlab{}.
\newblock \showarticletitle{GNNDelete: A general unlearning strategy for graph neural networks}. In \bibinfo{booktitle}{\emph{ICLR}}.
\newblock


\bibitem[Chenjia et~al\mbox{.}(2025)]%
        {Bai2025ONLINE}
\bibfield{author}{\bibinfo{person}{Bai Chenjia}, \bibinfo{person}{Zhang Yang}, \bibinfo{person}{Qiu Shuang}, \bibinfo{person}{Zhang Qiaosheng}, \bibinfo{person}{Xu Kang}, {and} \bibinfo{person}{Li Xuelong}.} \bibinfo{year}{2025}\natexlab{}.
\newblock \showarticletitle{ONLINE PREFERENCE ALIGNMENT FOR LANGUAGE MODELS VIA COUNT-BASED EXPLORATION}. In \bibinfo{booktitle}{\emph{ICLR}}.
\newblock


\bibitem[Chenliang et~al\mbox{.}(2025)]%
        {Li2025LEARNING}
\bibfield{author}{\bibinfo{person}{Li Chenliang}, \bibinfo{person}{Zeng Siliang}, \bibinfo{person}{Liao Zeyi}, \bibinfo{person}{Li Jiaxiang}, \bibinfo{person}{Kang Dongyeop}, \bibinfo{person}{Garcia Alfredo}, {and} \bibinfo{person}{Hong Mingyi}.} \bibinfo{year}{2025}\natexlab{}.
\newblock \showarticletitle{LEARNING REWARD AND POLICY JOINTLY FROM DEMONSTRATION AND PREFERENCE IMPROVES ALIGNMENT}. In \bibinfo{booktitle}{\emph{ICLR}}.
\newblock


\bibitem[Cong and Mahdavi(2023)]%
        {Cong2023GraphEditor}
\bibfield{author}{\bibinfo{person}{Weilin Cong} {and} \bibinfo{person}{Mehrdad Mahdavi}.} \bibinfo{year}{2023}\natexlab{}.
\newblock \showarticletitle{GraphEditor: An efficient graph representation learning and unlearning approach}.
\newblock


\bibitem[Dai et~al\mbox{.}(2024)]%
        {Dai2024}
\bibfield{author}{\bibinfo{person}{Juntao Dai}, \bibinfo{person}{Xuehai Pan}, \bibinfo{person}{Ruiyang Sun}, \bibinfo{person}{Jiaming Ji}, \bibinfo{person}{Xinbo Xu}, \bibinfo{person}{Mickel Liu}, \bibinfo{person}{Yizhou Wang}, {and} \bibinfo{person}{Yaodong Yang}.} \bibinfo{year}{2024}\natexlab{}.
\newblock \showarticletitle{SAFE RLHF: SAFE REINFORCEMENT LEARNING FROM HUMAN FEEDBACK}. In \bibinfo{booktitle}{\emph{ICLR}}.
\newblock


\bibitem[Derrick et~al\mbox{.}(2019)]%
        {Blakely2019Time}
\bibfield{author}{\bibinfo{person}{Blakely Derrick}, \bibinfo{person}{Lanchantin Jack}, {and} \bibinfo{person}{Qi Yanjun}.} \bibinfo{year}{2019}\natexlab{}.
\newblock \showarticletitle{Time and Space Complexity of Graph Convolutional Networks}.
\newblock  (\bibinfo{year}{2019}).
\newblock
\urldef\tempurl%
\url{https://api.semanticscholar.org/CorpusID:269411067}
\showURL{%
\tempurl}


\bibitem[Edward~J et~al\mbox{.}(2022)]%
        {Hu2022LoRA}
\bibfield{author}{\bibinfo{person}{Hu Edward~J}, \bibinfo{person}{shen yelong}, \bibinfo{person}{Wallis Phillip}, \bibinfo{person}{Allen-Zhu Zeyuan}, \bibinfo{person}{Li Yuanzhi}, \bibinfo{person}{Wang Shean}, \bibinfo{person}{Wang Lu}, {and} \bibinfo{person}{Weizhu Chen}.} \bibinfo{year}{2022}\natexlab{}.
\newblock \showarticletitle{LoRA: Low-Rank Adaptation of Large Language Models}. In \bibinfo{booktitle}{\emph{ICLR}}.
\newblock


\bibitem[Fan et~al\mbox{.}(2024)]%
        {Fan2024Simplicity}
\bibfield{author}{\bibinfo{person}{Chongyu Fan}, \bibinfo{person}{Jiancheng Liu}, \bibinfo{person}{Licong Lin}, \bibinfo{person}{Jinghan Jia}, \bibinfo{person}{Ruiqi Zhang}, \bibinfo{person}{Song Mei}, {and} \bibinfo{person}{Sijia Liu}.} \bibinfo{year}{2024}\natexlab{}.
\newblock \showarticletitle{Simplicity Prevails: Rethinking Negative Preference Optimization for LLM Unlearning}. In \bibinfo{booktitle}{\emph{NeurPIS}}.
\newblock


\bibitem[Fan et~al\mbox{.}(2025)]%
        {fan2025manta}
\bibfield{author}{\bibinfo{person}{Lei Fan}, \bibinfo{person}{Dongdong Fan}, \bibinfo{person}{Zhiguang Hu}, \bibinfo{person}{Yiwen Ding}, \bibinfo{person}{Donglin Di}, \bibinfo{person}{Kai Yi}, \bibinfo{person}{Maurice Pagnucco}, {and} \bibinfo{person}{Yang Song}.} \bibinfo{year}{2025}\natexlab{}.
\newblock \showarticletitle{Manta: A large-scale multi-view and visual-text anomaly detection dataset for tiny objects}. In \bibinfo{booktitle}{\emph{Proceedings of the Computer Vision and Pattern Recognition Conference}}. \bibinfo{pages}{25518--25527}.
\newblock


\bibitem[Gilmer et~al\mbox{.}(2017)]%
        {Gilmer2017NeuralMessagePassing}
\bibfield{author}{\bibinfo{person}{Justin Gilmer}, \bibinfo{person}{Samuel~S. Schoenholz}, \bibinfo{person}{Patrick~F. Riley}, \bibinfo{person}{Oriol Vinyals}, {and} \bibinfo{person}{George~E. Dahl}.} \bibinfo{year}{2017}\natexlab{}.
\newblock \showarticletitle{Neural message passing for quantum chemistry}. In \bibinfo{booktitle}{\emph{International Conference on Machine Learning}}. \bibinfo{publisher}{PMLR}, \bibinfo{pages}{1263--1272}.
\newblock


\bibitem[Haldar et~al\mbox{.}(2025)]%
        {Haldar2025LLM}
\bibfield{author}{\bibinfo{person}{Rajdeep Haldar}, \bibinfo{person}{Wang Ziyi}, \bibinfo{person}{Song Qifan}, \bibinfo{person}{Lin Guang}, {and} \bibinfo{person}{Xing Yue}.} \bibinfo{year}{2025}\natexlab{}.
\newblock \showarticletitle{LLM Safety Alignment is Divergence Estimation in Disguise}.
\newblock \bibinfo{journal}{\emph{arXiv preprint arXiv:2502.00657}} (\bibinfo{year}{2025}).
\newblock


\bibitem[Hamilton et~al\mbox{.}(2017)]%
        {Hamilton}
\bibfield{author}{\bibinfo{person}{William~L. Hamilton}, \bibinfo{person}{Rex Ying}, {and} \bibinfo{person}{Jure Leskovec}.} \bibinfo{year}{2017}\natexlab{}.
\newblock \showarticletitle{Inductive representation learning on large graphs}. In \bibinfo{booktitle}{\emph{NeurIPS}}.
\newblock


\bibitem[Hu et~al\mbox{.}(2020)]%
        {hu2024open}
\bibfield{author}{\bibinfo{person}{Weihua Hu}, \bibinfo{person}{Matthias Fey}, \bibinfo{person}{Marinka Zitnik}, \bibinfo{person}{Yuxiao Dong}, \bibinfo{person}{Hongyu Ren}, \bibinfo{person}{Bowen Liu}, \bibinfo{person}{Michele Catasta}, {and} \bibinfo{person}{Jure Leskovec}.} \bibinfo{year}{2020}\natexlab{}.
\newblock \showarticletitle{Open graph benchmark: datasets for machine learning on graphs}. In \bibinfo{booktitle}{\emph{NeurIPS}}.
\newblock


\bibitem[Ji et~al\mbox{.}(2024)]%
        {Ji2024}
\bibfield{author}{\bibinfo{person}{Jiaming Ji}, \bibinfo{person}{Donghai Hong}, \bibinfo{person}{Borong Zhang}, \bibinfo{person}{Boyuan Chen}, \bibinfo{person}{Josef Dai}, \bibinfo{person}{Boren Zheng}, \bibinfo{person}{Tianyi Qiu}, \bibinfo{person}{Boxun Li}, {and} \bibinfo{person}{Yaodong Yang}.} \bibinfo{year}{2024}\natexlab{}.
\newblock \showarticletitle{PKU-SafeRLHF: Towards Multi-Level Safety Alignment for LLMs with Human Preference}. In \bibinfo{booktitle}{\emph{NeurPIS}}.
\newblock


\bibitem[Junkang et~al\mbox{.}(2025)]%
        {Wu2025TOWARDS}
\bibfield{author}{\bibinfo{person}{Wu Junkang}, \bibinfo{person}{Xie Yuexiang}, \bibinfo{person}{Yang Zhengyi}, \bibinfo{person}{Wu Jiancan}, \bibinfo{person}{Chen Jiawei}, \bibinfo{person}{Gao Jinyang}, \bibinfo{person}{Ding Bolin}, \bibinfo{person}{Wang Xiang}, {and} \bibinfo{person}{He Xiangnan}.} \bibinfo{year}{2025}\natexlab{}.
\newblock \showarticletitle{TOWARDS ROBUST ALIGNMENT OF LANGUAGE MODELS: DISTRIBUTIONALLY ROBUSTIFYING DIRECT PREFERENCE OPTIMIZATION}. In \bibinfo{booktitle}{\emph{ICLR}}.
\newblock


\bibitem[Kipf and Max(2017)]%
        {Kipf}
\bibfield{author}{\bibinfo{person}{Thomas~N. Kipf} {and} \bibinfo{person}{Welling Max}.} \bibinfo{year}{2017}\natexlab{}.
\newblock \showarticletitle{Semi-supervised classification with graph convolutional networks}. In \bibinfo{booktitle}{\emph{ICLR}}.
\newblock


\bibitem[Li et~al\mbox{.}(2025a)]%
        {Li2023Aaron}
\bibfield{author}{\bibinfo{person}{Aaron~Jiaxun Li}, \bibinfo{person}{Satyapriya Krishna}, {and} \bibinfo{person}{Himabindu Lakkaraju}.} \bibinfo{year}{2025}\natexlab{a}.
\newblock \showarticletitle{Joint Reward and Policy Learning with Demonstrations and Human Feedback Improves Alignment}. In \bibinfo{booktitle}{\emph{ICLR}}.
\newblock


\bibitem[Li et~al\mbox{.}(2025b)]%
        {Li2023}
\bibfield{author}{\bibinfo{person}{Chenliang Li}, \bibinfo{person}{Siliang Zeng}, \bibinfo{person}{Zeyi Liao}, \bibinfo{person}{Jiaxiang Li}, \bibinfo{person}{Dongyeop Kang}, \bibinfo{person}{Alfredo Garcia}, {and} \bibinfo{person}{Hong Mingyi}.} \bibinfo{year}{2025}\natexlab{b}.
\newblock \showarticletitle{Joint Reward and Policy Learning with Demonstrations and Human Feedback Improves Alignment}. In \bibinfo{booktitle}{\emph{ICLR}}.
\newblock


\bibitem[Li et~al\mbox{.}(2024b)]%
        {Li2024}
\bibfield{author}{\bibinfo{person}{Jianing Li}, \bibinfo{person}{Chaoqun Yang}, \bibinfo{person}{Guanhua Ye}, {and} \bibinfo{person}{Quoc Viet~Hung Nguyen}.} \bibinfo{year}{2024}\natexlab{b}.
\newblock \showarticletitle{Graph neural networks with deep mutual learning for designing multi-modal recommendation systems}.
\newblock \bibinfo{journal}{\emph{Information Sciences}} (\bibinfo{year}{2024}).
\newblock


\bibitem[Li et~al\mbox{.}(2024a)]%
        {li2024fast}
\bibfield{author}{\bibinfo{person}{Weikai Li}, \bibinfo{person}{Zhiping Xiao}, \bibinfo{person}{Xiao Luo}, {and} \bibinfo{person}{Yizhou Sun}.} \bibinfo{year}{2024}\natexlab{a}.
\newblock \showarticletitle{Fast Inference of Removal-Based Node Influence}. In \bibinfo{booktitle}{\emph{Proceedings of the ACM Web Conference 2024 (WWW '24)}}. \bibinfo{publisher}{Association for Computing Machinery}, \bibinfo{address}{New York, NY, USA}.
\newblock


\bibitem[Li et~al\mbox{.}(2024c)]%
        {Li2024Towards}
\bibfield{author}{\bibinfo{person}{Xunkai Li}, \bibinfo{person}{Yulin Zhao}, \bibinfo{person}{Zhengyu Wu}, \bibinfo{person}{Wentao Zhang}, \bibinfo{person}{Rong-Hua Li}, {and} \bibinfo{person}{Guoren Wang}.} \bibinfo{year}{2024}\natexlab{c}.
\newblock \showarticletitle{Towards Effective and General Graph Unlearning via Mutual Evolution}. In \bibinfo{booktitle}{\emph{AAAI}}, Vol.~\bibinfo{volume}{38}. \bibinfo{pages}{13682--13690}.
\newblock


\bibitem[Liang et~al\mbox{.}(2024a)]%
        {Liang2024}
\bibfield{author}{\bibinfo{person}{Ke Liang}, \bibinfo{person}{Lingyuan Meng}, \bibinfo{person}{Meng Liu}, \bibinfo{person}{Yue Liu}, \bibinfo{person}{Wenxuan Tu}, \bibinfo{person}{Siwei Wang}, \bibinfo{person}{Sihang Zhou}, {and} \bibinfo{person}{Xinwang Liu}.} \bibinfo{year}{2024}\natexlab{a}.
\newblock \showarticletitle{A survey of knowledge graph reasoning on graph types: Static, dynamic, and multi-modal}.
\newblock \bibinfo{journal}{\emph{IEEE Transactions on Pattern Analysis and Machine Intelligence}} (\bibinfo{year}{2024}).
\newblock


\bibitem[Liang et~al\mbox{.}(2024b)]%
        {Liang2024:56}
\bibfield{author}{\bibinfo{person}{Wanying Liang}, \bibinfo{person}{Pasquale~De Meo}, \bibinfo{person}{Yong Tang}, {and} \bibinfo{person}{Jia Zhu}.} \bibinfo{year}{2024}\natexlab{b}.
\newblock \showarticletitle{A survey of multi-modal knowledge graphs: Technologies and trends}.
\newblock \bibinfo{journal}{\emph{Comput. Surveys}} \bibinfo{volume}{56}, \bibinfo{number}{11} (\bibinfo{year}{2024}), \bibinfo{pages}{1--41}.
\newblock


\bibitem[Liu et~al\mbox{.}(2022)]%
        {Liu2022TheRight}
\bibfield{author}{\bibinfo{person}{Yi Liu}, \bibinfo{person}{Lei Xu}, \bibinfo{person}{Xingliang Yuan}, \bibinfo{person}{Cong Wang}, {and} \bibinfo{person}{Bo Li}.} \bibinfo{year}{2022}\natexlab{}.
\newblock \showarticletitle{The Right to be Forgotten in Federated Learning: An Efficient Realization with Rapid Retraining}. In \bibinfo{booktitle}{\emph{IEEE Conference on Computer Communications}}.
\newblock
\newblock
\shownote{2022b}.


\bibitem[Liu et~al\mbox{.}(2021)]%
        {Liu2021tail}
\bibfield{author}{\bibinfo{person}{Zemin Liu}, \bibinfo{person}{Trung-Kien Nguyen}, {and} \bibinfo{person}{Yuan Fang}.} \bibinfo{year}{2021}\natexlab{}.
\newblock \showarticletitle{Tail-GNN: Tail-Node Graph Neural Networks}. In \bibinfo{booktitle}{\emph{Proceedings of the 27th ACM SIGKDD Conference on Knowledge Discovery and Data Mining}}. \bibinfo{publisher}{Association for Computing Machinery}, \bibinfo{pages}{1109--1119}.
\newblock


\bibitem[Olatunji et~al\mbox{.}(2021)]%
        {Olatunji2021MembershipInferenceGNN}
\bibfield{author}{\bibinfo{person}{Iyiola~E Olatunji}, \bibinfo{person}{Wolfgang Nejdl}, {and} \bibinfo{person}{Megha Khosla}.} \bibinfo{year}{2021}\natexlab{}.
\newblock \showarticletitle{Membership inference attack on graph neural networks}. In \bibinfo{booktitle}{\emph{Proceedings of the IEEE International Conference on Trust, Privacy and Security in Intelligent Systems and Applications}}.
\newblock


\bibitem[Pratyush et~al\mbox{.}(2024)]%
        {Maini2024TOFU}
\bibfield{author}{\bibinfo{person}{Maini Pratyush}, \bibinfo{person}{Feng Zhili}, \bibinfo{person}{Schwarzschild Avi}, \bibinfo{person}{Lipton Zachary~C.}, {and} \bibinfo{person}{Zico~Kolter J.}} \bibinfo{year}{2024}\natexlab{}.
\newblock \showarticletitle{TOFU: A Task of Fictitious Unlearning for LLMs}. In \bibinfo{booktitle}{\emph{NeurIPS}}.
\newblock


\bibitem[Rafailov et~al\mbox{.}(2023)]%
        {Rafailov2023}
\bibfield{author}{\bibinfo{person}{Rafael Rafailov}, \bibinfo{person}{Archit Sharma}, \bibinfo{person}{Eric Mitchell}, \bibinfo{person}{Stefano Ermon}, \bibinfo{person}{Christopher~D. Manning}, {and} \bibinfo{person}{Chelsea Finn}.} \bibinfo{year}{2023}\natexlab{}.
\newblock \showarticletitle{Direct Preference Optimization: Your Language Model is Secretly a Reward Model}. In \bibinfo{booktitle}{\emph{NeurPIS}}.
\newblock


\bibitem[Rong et~al\mbox{.}(2023)]%
        {Rong2023EfficientGNN}
\bibfield{author}{\bibinfo{person}{Yao Rong}, \bibinfo{person}{Guanchu Wang}, \bibinfo{person}{Qizhang Feng}, \bibinfo{person}{Ninghao Liu}, \bibinfo{person}{Zirui Liu}, \bibinfo{person}{Enkelejda Kasneci}, {and} \bibinfo{person}{Xia Hu}.} \bibinfo{year}{2023}\natexlab{}.
\newblock \bibinfo{title}{Efficient GNN Explanation via Learning Removal-based Attribution}.
\newblock
\showeprint[arxiv]{2306.05760}~[cs.LG]


\bibitem[Sablayrolles et~al\mbox{.}(2019)]%
        {Sablayrolles2019WhiteboxBlackbox}
\bibfield{author}{\bibinfo{person}{Alexandre Sablayrolles}, \bibinfo{person}{Matthijs Douze}, \bibinfo{person}{Yann Ollivier}, \bibinfo{person}{Cordelia Schmid}, {and} \bibinfo{person}{Hervé Jégou}.} \bibinfo{year}{2019}\natexlab{}.
\newblock \showarticletitle{White-box vs black-box: Bayes optimal strategies for membership inference}. In \bibinfo{booktitle}{\emph{Proceedings of the International Conference on Machine Learning}}.
\newblock


\bibitem[Shi et~al\mbox{.}(2024)]%
        {Shi2024}
\bibfield{author}{\bibinfo{person}{Lei Shi}, \bibinfo{person}{Jiapeng Yang}, \bibinfo{person}{Pengtao Lv}, \bibinfo{person}{Lu Yuan}, \bibinfo{person}{Feifei Kou}, \bibinfo{person}{Jia Luo}, {and} \bibinfo{person}{Mingying Xu}.} \bibinfo{year}{2024}\natexlab{}.
\newblock \showarticletitle{Self-derived Knowledge Graph Contrastive Learning for Recommendation}. In \bibinfo{booktitle}{\emph{Proceedings of the 31st ACM International Conference on Multimedia (MM’23)}}. \bibinfo{publisher}{Association for Computing Machinery}, \bibinfo{address}{New York, NY, USA}, \bibinfo{pages}{7571–7580}.
\newblock


\bibitem[Shi et~al\mbox{.}(2025)]%
        {Shi2025MUSE}
\bibfield{author}{\bibinfo{person}{Weijia Shi}, \bibinfo{person}{Jaechan Lee}, \bibinfo{person}{Yangsibo Huang}, \bibinfo{person}{Sadhika Malladi}, \bibinfo{person}{Jieyu Zhao}, \bibinfo{person}{Ari Holtzman}, \bibinfo{person}{Daogao Liu}, \bibinfo{person}{Luke Zettlemoyer}, \bibinfo{person}{Noah~A. Smith}, {and} \bibinfo{person}{Chiyuan Zhang}.} \bibinfo{year}{2025}\natexlab{}.
\newblock \showarticletitle{MUSE: Machine Unlearning Six-Way Evaluation for Language Models}. In \bibinfo{booktitle}{\emph{ICLR}}.
\newblock


\bibitem[Stepin et~al\mbox{.}(2021)]%
        {Stepin2021Survey}
\bibfield{author}{\bibinfo{person}{Ilia Stepin}, \bibinfo{person}{Jose~M. Alonso}, \bibinfo{person}{Alejandro Catala}, {and} \bibinfo{person}{Mart{\'i}n Pereira-Fari{\~n}a}.} \bibinfo{year}{2021}\natexlab{}.
\newblock \showarticletitle{A Survey of Contrastive and Counterfactual Explanation Generation Methods for Explainable Artificial Intelligence}.
\newblock \bibinfo{journal}{\emph{IEEE Access}}  \bibinfo{volume}{9} (\bibinfo{year}{2021}), \bibinfo{pages}{11974--12001}.
\newblock
\href{https://doi.org/10.1109/ACCESS.2021.3051315}{doi:\nolinkurl{10.1109/ACCESS.2021.3051315}}


\bibitem[Su et~al\mbox{.}(2021a)]%
        {su2021prioritized}
\bibfield{author}{\bibinfo{person}{Xiu Su}, \bibinfo{person}{Tao Huang}, \bibinfo{person}{Yanxi Li}, \bibinfo{person}{Shan You}, \bibinfo{person}{Fei Wang}, \bibinfo{person}{Chen Qian}, \bibinfo{person}{Changshui Zhang}, {and} \bibinfo{person}{Chang Xu}.} \bibinfo{year}{2021}\natexlab{a}.
\newblock \showarticletitle{Prioritized architecture sampling with monto-carlo tree search}. In \bibinfo{booktitle}{\emph{Proceedings of the IEEE/CVF Conference on Computer Vision and Pattern Recognition}}. \bibinfo{pages}{10968--10977}.
\newblock


\bibitem[Su et~al\mbox{.}(2021b)]%
        {su2021locally}
\bibfield{author}{\bibinfo{person}{Xiu Su}, \bibinfo{person}{Shan You}, \bibinfo{person}{Tao Huang}, \bibinfo{person}{Fei Wang}, \bibinfo{person}{Chen Qian}, \bibinfo{person}{Changshui Zhang}, {and} \bibinfo{person}{Chang Xu}.} \bibinfo{year}{2021}\natexlab{b}.
\newblock \showarticletitle{Locally free weight sharing for network width search}.
\newblock \bibinfo{journal}{\emph{arXiv preprint arXiv:2102.05258}} (\bibinfo{year}{2021}).
\newblock


\bibitem[Su et~al\mbox{.}(2021c)]%
        {su2021bcnet}
\bibfield{author}{\bibinfo{person}{Xiu Su}, \bibinfo{person}{Shan You}, \bibinfo{person}{Fei Wang}, \bibinfo{person}{Chen Qian}, \bibinfo{person}{Changshui Zhang}, {and} \bibinfo{person}{Chang Xu}.} \bibinfo{year}{2021}\natexlab{c}.
\newblock \showarticletitle{Bcnet: Searching for network width with bilaterally coupled network}. In \bibinfo{booktitle}{\emph{Proceedings of the IEEE/CVF Conference on Computer Vision and Pattern Recognition}}. \bibinfo{pages}{2175--2184}.
\newblock


\bibitem[Su et~al\mbox{.}(2022a)]%
        {su2022searching}
\bibfield{author}{\bibinfo{person}{Xiu Su}, \bibinfo{person}{Shan You}, \bibinfo{person}{Jiyang Xie}, \bibinfo{person}{Fei Wang}, \bibinfo{person}{Chen Qian}, \bibinfo{person}{Changshui Zhang}, {and} \bibinfo{person}{Chang Xu}.} \bibinfo{year}{2022}\natexlab{a}.
\newblock \showarticletitle{Searching for network width with bilaterally coupled network}.
\newblock \bibinfo{journal}{\emph{IEEE Transactions on Pattern Analysis and Machine Intelligence}} \bibinfo{volume}{45}, \bibinfo{number}{7} (\bibinfo{year}{2022}), \bibinfo{pages}{8936--8953}.
\newblock


\bibitem[Su et~al\mbox{.}(2022b)]%
        {su2022vitas}
\bibfield{author}{\bibinfo{person}{Xiu Su}, \bibinfo{person}{Shan You}, \bibinfo{person}{Jiyang Xie}, \bibinfo{person}{Mingkai Zheng}, \bibinfo{person}{Fei Wang}, \bibinfo{person}{Chen Qian}, \bibinfo{person}{Changshui Zhang}, \bibinfo{person}{Xiaogang Wang}, {and} \bibinfo{person}{Chang Xu}.} \bibinfo{year}{2022}\natexlab{b}.
\newblock \showarticletitle{ViTAS: Vision transformer architecture search}. In \bibinfo{booktitle}{\emph{European Conference on Computer Vision}}. Springer, \bibinfo{pages}{139--157}.
\newblock


\bibitem[Su et~al\mbox{.}(2021d)]%
        {su2021k}
\bibfield{author}{\bibinfo{person}{Xiu Su}, \bibinfo{person}{Shan You}, \bibinfo{person}{Mingkai Zheng}, \bibinfo{person}{Fei Wang}, \bibinfo{person}{Chen Qian}, \bibinfo{person}{Changshui Zhang}, {and} \bibinfo{person}{Chang Xu}.} \bibinfo{year}{2021}\natexlab{d}.
\newblock \showarticletitle{K-shot nas: Learnable weight-sharing for nas with k-shot supernets}. In \bibinfo{booktitle}{\emph{International Conference on Machine Learning}}. PMLR, \bibinfo{pages}{9880--9890}.
\newblock


\bibitem[Sungmin et~al\mbox{.}(2025)]%
        {Cha2025TOWARDS}
\bibfield{author}{\bibinfo{person}{Cha Sungmin}, \bibinfo{person}{Cho Sungjun}, \bibinfo{person}{Hwang Dasol}, {and} \bibinfo{person}{Lee Moontae}.} \bibinfo{year}{2025}\natexlab{}.
\newblock \showarticletitle{TOWARDS ROBUST AND PARAMETER-EFFICIENT KNOWLEDGE UNLEARNING FOR LLMS}. In \bibinfo{booktitle}{\emph{ICLR}}.
\newblock


\bibitem[Tan et~al\mbox{.}(2024)]%
        {Tan2024Unlink}
\bibfield{author}{\bibinfo{person}{Jiajun Tan}, \bibinfo{person}{Fei Sun}, \bibinfo{person}{Ruichen Qiu}, \bibinfo{person}{Du Su}, {and} \bibinfo{person}{Huawei Shen}.} \bibinfo{year}{2024}\natexlab{}.
\newblock \showarticletitle{Unlink to Unlearn: Simplifying Edge Unlearning in GNNs}. In \bibinfo{booktitle}{\emph{Companion Proceedings of the ACM Web Conference 2024 (WWW '24)}}. \bibinfo{publisher}{Association for Computing Machinery}, \bibinfo{address}{New York, NY, USA}, \bibinfo{pages}{489--492}.
\newblock


\bibitem[Tang et~al\mbox{.}(2020)]%
        {Tang2020investigating}
\bibfield{author}{\bibinfo{person}{Xianfeng Tang}, \bibinfo{person}{Huaxiu Yao}, \bibinfo{person}{Yiwei Sun}, \bibinfo{person}{Yiqi Wang}, \bibinfo{person}{Jiliang Tang}, \bibinfo{person}{Charu Aggarwal}, \bibinfo{person}{Prasenjit Mitra}, {and} \bibinfo{person}{Suhang Wang}.} \bibinfo{year}{2020}\natexlab{}.
\newblock \showarticletitle{Investigating and Mitigating Degree-Related Biases in Graph Convolutional Networks}. In \bibinfo{booktitle}{\emph{Proceedings of the 29th ACM International Conference on Information and Knowledge Management}}. \bibinfo{publisher}{Association for Computing Machinery}, \bibinfo{pages}{1435--1444}.
\newblock


\bibitem[Voigt and dem Bussche(2017)]%
        {Voigt2017}
\bibfield{author}{\bibinfo{person}{Paul Voigt} {and} \bibinfo{person}{Axel~Von dem Bussche}.} \bibinfo{year}{2017}\natexlab{}.
\newblock \bibinfo{booktitle}{\emph{The EU General Data Protection Regulation (GDPR): A Practical Guide} (\bibinfo{edition}{1st} ed.)}.
\newblock \bibinfo{publisher}{Springer International Publishing}, \bibinfo{address}{Cham}.
\newblock
\showISBNx{978-3-319-57276-7}


\bibitem[Wang et~al\mbox{.}(2025a)]%
        {Wang2025}
\bibfield{author}{\bibinfo{person}{Qizhou Wang}, \bibinfo{person}{Bo Han}, \bibinfo{person}{Puning Yang}, \bibinfo{person}{Jianing Zhu}, \bibinfo{person}{Tongliang Liu}, {and} \bibinfo{person}{Masashi Sugiyama}.} \bibinfo{year}{2025}\natexlab{a}.
\newblock \showarticletitle{TOWARDS EFFECTIVE EVALUATIONS AND COMPARISONS FOR LLM UNLEARNING METHODS}. In \bibinfo{booktitle}{\emph{ICLR}}.
\newblock


\bibitem[Wang et~al\mbox{.}(2025c)]%
        {Wang2025RETHINKING}
\bibfield{author}{\bibinfo{person}{Qizhou Wang}, \bibinfo{person}{Jin~Peng Zhou}, \bibinfo{person}{Zhanke Zhou}, \bibinfo{person}{Saebyeol Shin}, \bibinfo{person}{Bo Han}, {and} \bibinfo{person}{Kilian~Q. Weinberger}.} \bibinfo{year}{2025}\natexlab{c}.
\newblock \showarticletitle{RETHINKING LLM UNLEARNING OBJECTIVES: A GRADIENT PERSPECTIVE AND GO BEYOND}. In \bibinfo{booktitle}{\emph{ICLR}}.
\newblock


\bibitem[Wang et~al\mbox{.}(2025b)]%
        {Wang2025TOWARDS}
\bibfield{author}{\bibinfo{person}{Yu Wang}, \bibinfo{person}{Tong Zhao}, \bibinfo{person}{Yuying Zhao}, \bibinfo{person}{Yunchao Liu}, \bibinfo{person}{Xueqi Cheng}, \bibinfo{person}{Neil Shah}, {and} \bibinfo{person}{Tyler Derr}.} \bibinfo{year}{2025}\natexlab{b}.
\newblock \showarticletitle{TOWARDS EFFECTIVE EVALUATIONS AND COMPARISONS FOR LLM UNLEARNING METHODS}. In \bibinfo{booktitle}{\emph{ICLR}}.
\newblock


\bibitem[Wu et~al\mbox{.}(2023)]%
        {Wu2023GIF}
\bibfield{author}{\bibinfo{person}{Jiancan Wu}, \bibinfo{person}{Yi Yang}, \bibinfo{person}{Yuchun Qian}, \bibinfo{person}{Yongduo Sui}, \bibinfo{person}{Xiang Wang}, {and} \bibinfo{person}{Xiangnan He}.} \bibinfo{year}{2023}\natexlab{}.
\newblock \showarticletitle{GIF: A General Graph Unlearning Strategy via Influence Function}. In \bibinfo{booktitle}{\emph{Proceedings of the ACM Web Conference 2023 (WWW '23)}}. \bibinfo{publisher}{Association for Computing Machinery}, \bibinfo{address}{New York, NY, USA}, \bibinfo{pages}{651--661}.
\newblock


\bibitem[Wu et~al\mbox{.}(2020)]%
        {Wu2020GNN}
\bibfield{author}{\bibinfo{person}{Zonghan Wu}, \bibinfo{person}{Shirui Pan}, \bibinfo{person}{Fengwen Chen}, \bibinfo{person}{Guodong Long}, \bibinfo{person}{Chengqi Zhang}, {and} \bibinfo{person}{S.~Yu Philip}.} \bibinfo{year}{2020}\natexlab{}.
\newblock \showarticletitle{A comprehensive survey on graph neural networks}.
\newblock \bibinfo{journal}{\emph{IEEE Transactions on Neural Networks and Learning Systems}} \bibinfo{volume}{32}, \bibinfo{number}{1} (\bibinfo{year}{2020}), \bibinfo{pages}{4--24}.
\newblock


\bibitem[Yan et~al\mbox{.}(2025)]%
        {Scholten2025PROBABILISTIC}
\bibfield{author}{\bibinfo{person}{Scholten Yan}, \bibinfo{person}{Gunnemann Stephan}, {and} \bibinfo{person}{Schwinn Leo}.} \bibinfo{year}{2025}\natexlab{}.
\newblock \showarticletitle{A PROBABILISTIC PERSPECTIVE ON UNLEARNING AND ALIGNMENT FOR LARGE LANGUAGE MODELS}. In \bibinfo{booktitle}{\emph{ICLR}}.
\newblock


\bibitem[Yao et~al\mbox{.}(2024)]%
        {Yao2024Large}
\bibfield{author}{\bibinfo{person}{Yuanshun Yao}, \bibinfo{person}{Xiaojun Xu}, {and} \bibinfo{person}{Yang Liu}.} \bibinfo{year}{2024}\natexlab{}.
\newblock \showarticletitle{Large Language Model Unlearning}. In \bibinfo{booktitle}{\emph{ICLR}}.
\newblock


\bibitem[Yeom et~al\mbox{.}(2019)]%
        {Yeom2019PrivacyRisk}
\bibfield{author}{\bibinfo{person}{Samuel Yeom}, \bibinfo{person}{Irene Giacomelli}, \bibinfo{person}{Matt Fredrikson}, {and} \bibinfo{person}{Somesh Jha}.} \bibinfo{year}{2019}\natexlab{}.
\newblock \showarticletitle{Privacy risk in machine learning: Analyzing the connection to overfitting}. In \bibinfo{booktitle}{\emph{ACM SIGSAC Conference on Computer and Communications Security}}.
\newblock


\bibitem[Yuan et~al\mbox{.}(2020)]%
        {Yuan2020XGNN}
\bibfield{author}{\bibinfo{person}{Hao Yuan}, \bibinfo{person}{Jiliang Tang}, \bibinfo{person}{Xia Hu}, {and} \bibinfo{person}{Shuiwang Ji}.} \bibinfo{year}{2020}\natexlab{}.
\newblock \bibinfo{title}{XGNN: Towards Model-Level Explanations of Graph Neural Networks}.
\newblock
\showeprint[arxiv]{2006.02587}~[cs.LG]


\bibitem[Yuan et~al\mbox{.}(2021)]%
        {Yuan2021ExplainabilitySubgraphExplorations}
\bibfield{author}{\bibinfo{person}{Hao Yuan}, \bibinfo{person}{Haiyang Yu}, \bibinfo{person}{Jie Wang}, \bibinfo{person}{Kang Li}, {and} \bibinfo{person}{Shuiwang Ji}.} \bibinfo{year}{2021}\natexlab{}.
\newblock \bibinfo{title}{On Explainability of Graph Neural Networks via Subgraph Explorations}.
\newblock
\showeprint[arxiv]{2102.05152}~[cs.LG]


\bibitem[Yuan et~al\mbox{.}(2025)]%
        {Yuan2025}
\bibfield{author}{\bibinfo{person}{Xiaojian Yuan}, \bibinfo{person}{Tianyu Pang}, \bibinfo{person}{Chao Du}, \bibinfo{person}{Kejiang Chen}, \bibinfo{person}{Weiming Zhang}, {and} \bibinfo{person}{Min Lin}.} \bibinfo{year}{2025}\natexlab{}.
\newblock \showarticletitle{A CLOSER LOOK AT MACHINE UNLEARNING FOR LARGE LANGUAGE MODELS}. In \bibinfo{booktitle}{\emph{ICLR}}.
\newblock


\bibitem[Zhang et~al\mbox{.}(2024)]%
        {Zhang2024}
\bibfield{author}{\bibinfo{person}{Ruiqi Zhang}, \bibinfo{person}{Licong Lin}, \bibinfo{person}{Yu Bai}, {and} \bibinfo{person}{Song Mei}.} \bibinfo{year}{2024}\natexlab{}.
\newblock \showarticletitle{Negative Preference Optimization: From Catastrophic Collapse to Effective Unlearning}. In \bibinfo{booktitle}{\emph{CoLM}}.
\newblock


\bibitem[Zhao et~al\mbox{.}(2025)]%
        {Zhao2025DO}
\bibfield{author}{\bibinfo{person}{Siyan Zhao}, \bibinfo{person}{Mingyi Hong}, \bibinfo{person}{Yang Liu}, \bibinfo{person}{Devamanyu Hazarika}, {and} \bibinfo{person}{Kaixiang Lin}.} \bibinfo{year}{2025}\natexlab{}.
\newblock \showarticletitle{DO LLMS RECOGNIZE YOUR PREFERENCES? EVALUATING PERSONALIZED PREFERENCE FOLLOWING IN LLMS}. In \bibinfo{booktitle}{\emph{ICLR}}.
\newblock


\bibitem[Zhu et~al\mbox{.}(2025)]%
        {zhu2025interpretable}
\bibfield{author}{\bibinfo{person}{Zhijie Zhu}, \bibinfo{person}{Lei Fan}, \bibinfo{person}{Maurice Pagnucco}, {and} \bibinfo{person}{Yang Song}.} \bibinfo{year}{2025}\natexlab{}.
\newblock \showarticletitle{Interpretable Image Classification via Non-parametric Part Prototype Learning}. In \bibinfo{booktitle}{\emph{Proceedings of the Computer Vision and Pattern Recognition Conference}}. \bibinfo{pages}{9762--9771}.
\newblock


\end{thebibliography}

\appendix

\section{Appendices}
\subsection{Related Work}
\textbf{Graph Unlearning}. Retraining\cite{Liu2022TheRight} refers to train the model from scratch to unlearn specific edges rather than fine-tuning and is inefficient. GraphEraser\cite{Chen2022GraphUnlearning} attempts to achieve graph unlearning by employing graph partitioning and efficient retraining, but it support only node deletion. GraphEditor\cite{Cong2023GraphEditor} provides a closed-form solution for linear GNNs to guarantee information deletion, and additional fine-tuning can improve model utility. However, GraphEditor\cite{Cong2023GraphEditor} is not designed for graph-structured data, which is only applicable to linear structures. GIF\cite{Wu2023GIF} accurately estimates parameter changes by designing influence functions to directly modify the parameters for edge unlearning, but this performance on forget set is poor and can not achieve true unlearning. GNNDelete\cite{Cheng2023} achieves unlearning by approximating the representation of edges to be forgotten to those that did not exist in the pretrained model, while keeping the neighbors' representations minimally changed. However, GNNDelete\cite{Cheng2023} is infeasible to distinguish the representation distance between the forgotten and the retained data, leading to poor robustness for delete ratio. MEGU\cite{Li2024Towards} propose a new mutual evolution paradigm that simultaneously evolves the utility and forget capacities of graph unlearning, which unlearning by gradient ascent with rapid divergence speed. Compared to GNNDelete\cite{Cheng2023}, UtU\cite{Tan2024Unlink} only uses the graph after edge deletion for a single inference. However, these models makes the utility vulnerable during the unlearning process due to the rapid divergence speed of gradient ascent, especially MEGU. In this work, we aim to improve the robustness of the model utility to the unlearning process.

\textbf{Large Language Model Unlearning}. Gradient Ascent\cite{Yao2024Large} utilize fine-tuning to minimize correct predictions on the forget set by modifying the cross-entropy loss. NPO\cite{Zhang2024} adjusts offline DPO\cite{Rafailov2023} to reduce the likelihood of the forget set, avoiding the complexity of learning a reward function like RLHF\cite{Dai2024}. SimNPO\cite{Fan2024Simplicity} propose a simple yet effective unlearning optimization framework to remove the reliance on a reference model. To address utility preservation, regularized optimization\cite{Maini2024TOFU,Shi2025MUSE,Yuan2025} combines unlearning efficacy with model utility loss, like Gradient Descent loss and KL-Loss. Despite various studies on LLM Unlearning, our study reveals that existing unlearning methods with regularization struggle with handling graph-structure data due to tight coupling between data entities. We propose a simple yet effective solution to improve utility robustness for graph unlearning.

\subsection{Overall Performance on All Datasets}
\label{appendix:overall:performance}

\begin{table}[h]
\vspace{-10pt}
\caption{Comparison results of our model with advanced fine-tuning methods on dataset PubMed.}
\label{fig:overall-performance:PubMed}
\vspace{-5pt}
\begin{center}
\begin{small}
\setlength{\tabcolsep}{6pt} 
\begin{tabular}{lcccccc}
\toprule
Model  & RT-AUC & FT-AUC & MI Ratio & $p_f$ & $p_r$ & $\frac{pr}{p_f}$ \\
\midrule
GIF         & 0.9643       & 0.4699     & 1.05       & 0.9302       & 0.9045    & 0.97              \\
GNNDelete   & 0.9610       & 0.9762     & 1.65       & 0.5919       & 0.8692    & 1.46              \\
UtU         & 0.9643       & 0.4585     & 1.05       & 0.9288       & 0.9030    & 0.97              \\
\textbf{INPO-S}         & 0.9668       & 0.9834       & 1.74       & 0.5639      & 0.8717   & 1.55        \\
\bottomrule
\end{tabular}
\end{small}
\end{center}
\end{table}

\begin{table}[h]
\vspace{-10pt}
\caption{Comparison results of our model with advanced fine-tuning methods on dataset CS.}
\label{fig:overall-performance:CS}
\vspace{-5pt}
\begin{center}
\begin{small}
\setlength{\tabcolsep}{6pt} 
\begin{tabular}{lcccccc}
\toprule
Model  & RT-AUC & FT-AUC & MI Ratio & $p_f$ & $p_r$ & $\frac{pr}{p_f}$ \\
\midrule
GIF         & 0.9621       & 0.9129     & 1.06       & 0.9240       & 0.9021    & 0.97              \\
GNNDelete   & 0.9515       & 0.9682     & 1.68       & 0.5805       & 0.8424    & 1.45              \\
UtU         & 0.9626       & 0.5233     & 1.06       & 0.9246       & 0.9027    & 0.97               \\
\textbf{INPO-S}         & 0.9525       & 0.9791       & 1.80       & 0.5423     & 0.8608   & 1.59        \\
\bottomrule
\end{tabular}
\end{small}
\end{center}
\end{table}

\begin{table}[h]
\vspace{-10pt}
\caption{Comparison results of our model with advanced fine-tuning methods on dataset OGB-Collab.}
\label{fig:overall-performance:Collab}
\vspace{-5pt}
\begin{center}
\begin{small}
\setlength{\tabcolsep}{6pt} 
\begin{tabular}{lcccccc}
\toprule
Model  & RT-AUC & FT-AUC & MI Ratio & $p_f$ & $p_r$ & $\frac{pr}{p_f}$ \\
\midrule
GIF         & 0.9824       & 0.4837     & 1.01       & 0.9665       & 0.9600    & 0.99              \\
GNNDelete   & 0.9850       & 0.7230     & 1.45       & 0.6714       & 0.8527    & 1.27               \\
UtU         & 0.9852       & 0.5013     & 1.04       & 0.9401       & 0.9340    & 0.99              \\
\textbf{INPO-S}         & 0.9827       & 0.7396       & 1.56       & 0.6299      & 0.8713   & 1.38        \\
\bottomrule
\end{tabular}
\end{small}
\end{center}
\end{table}

\textbf{The effectiveness of IMPNN}. As shown in Figure \ref{fig:IMPNN:wo}, IMPNN reduces the adaptive coefficient \(S_{\theta}(x,y)\) leading to a slower divergence speed, thus minimizing the impact on the model utility.
\label{IMPNN:wo}

\subsection{Hyper-parameters Setting}
We list all hyper-parameters setting to reproduce our experiments.

\begin{table}[h]
\vspace{-10pt}
\caption{Hyper-parameters Setting on All Datasets.}
\label{table:Hyper-parameters}
\vspace{-5pt}
\begin{center}
\begin{small}
\setlength{\tabcolsep}{6pt} 
\begin{tabular}{lcccccc}
\toprule
Hyper-parameter  & $\beta$ & epoch & lr & $\lambda_1$ & NI & $\lambda_3$ \\
\midrule
INPO-S         & 2       & 100     & 1e-3       & 1       & 1    & 0.2    \\     
\bottomrule
\end{tabular}
\end{small}
\end{center}
\end{table}

\begin{figure}[t]
  \centering
  \includegraphics[width=1\linewidth]{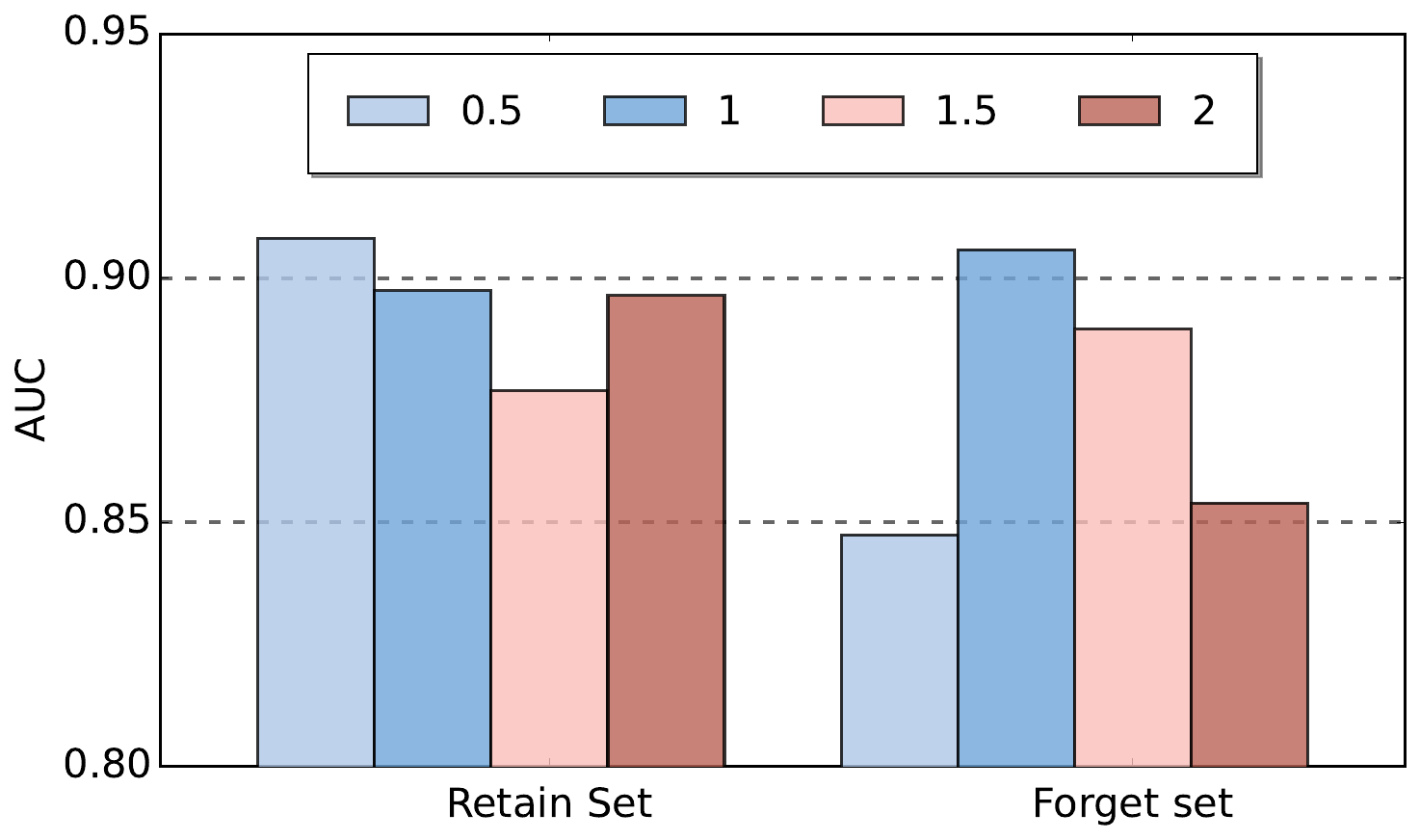}
  \caption{AUC Performance on the retain set and the forget set at different \(\lambda_1\).}
  \label{fig:sens:lambda1}
\end{figure}

\begin{figure}[t]
  \centering
  \includegraphics[width=1\linewidth]{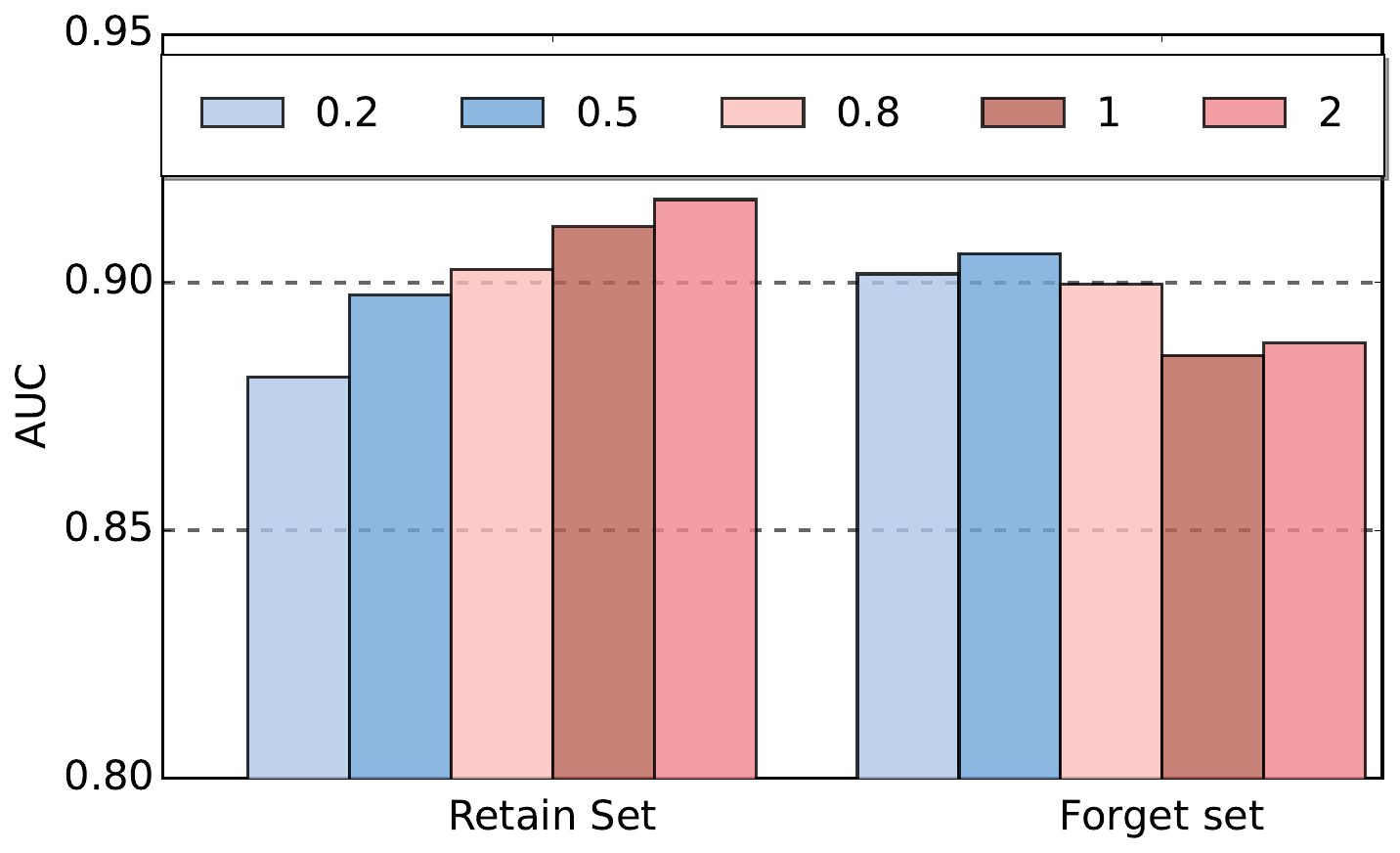}
  \caption{AUC Performance on the retain set and the forget set at different \(\lambda_3\).}
  \label{fig:sens:lambda3}
\end{figure}

\begin{figure}[t]
    \centering
    \begin{subfigure}[b]{0.23\textwidth}
        \includegraphics[width=\textwidth]{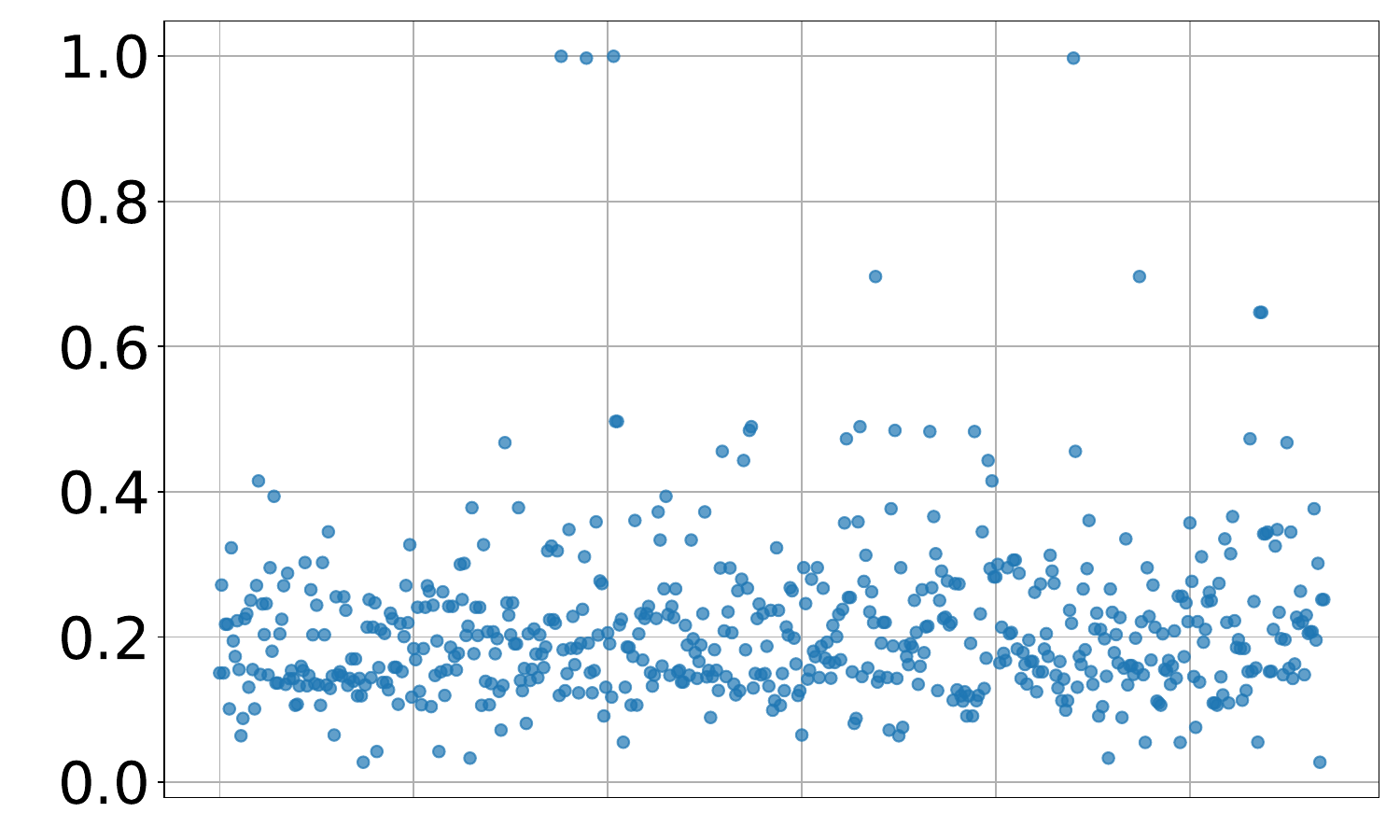}
        \caption{\(S_\theta(x, y)\) for MPNN.}
    \end{subfigure}
    \hfill
    \begin{subfigure}[b]{0.23\textwidth}
        \includegraphics[width=\textwidth]{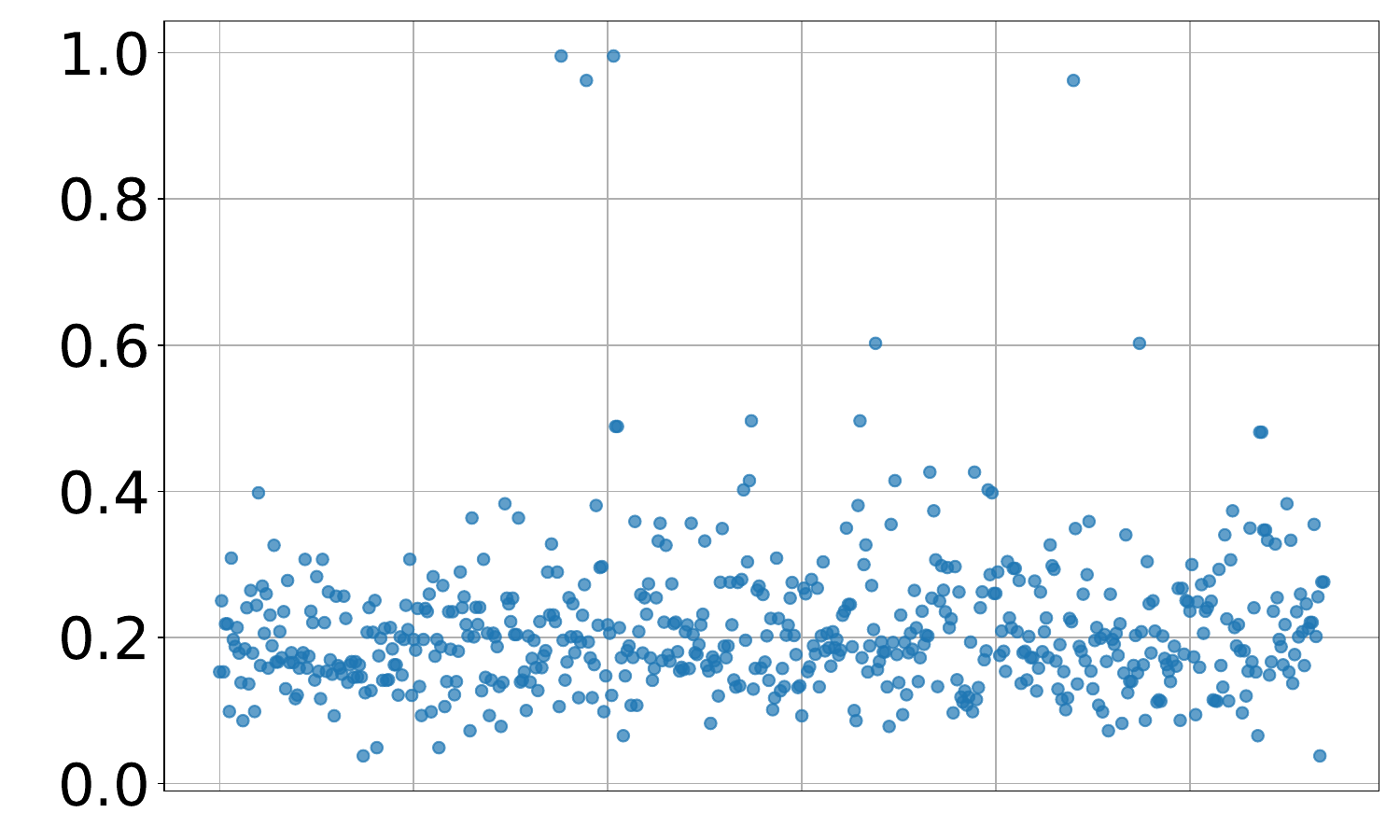}
        \caption{\(S_\theta(x, y)\) for IMPNN. }
    \end{subfigure}
    \caption{The adaptive coefficient \(S_\theta(x, y)\) at epoch 250 for MPNN and IMPNN. A data point represents an edge to be forgotten.}
    \label{fig:IMPNN:wo}
\end{figure}

\subsection{Proof of Our Model's Effectiveness}
\label{proof:effect:inpo}
In this section, we theoretically prove the effectiveness of the proposed method.

Graph Edge Unlearning is actually to perform separation of edge representations between the forget set and the retain set. 
The variational form of TV-Divergencen is:
\begin{equation}
\mathbb{D}_{TV}(\mathcal{P} || \mathcal{Q}) = \sup\limits_{f:|f| \leq 1/2} \mathbb{E}_{v\in\mathcal{P}}f(v) - \mathbb{E}_{v\in\mathcal{Q}}f(v),
\end{equation}
where \(f\) an arbitrary function.

The variational form of KL-Divergencen is:
\begin{equation}
\mathbb{D}_{KL}(\mathcal{P} || \mathcal{Q}) = \sup\limits_{f} \mathbb{E}_{v\in\mathcal{P}}f(v) - log(\mathbb{E}_{v\in\mathcal{Q}}e^{f(v)}).
\end{equation}
Let \(f=r_{\theta}(y \mid x) = \beta \cdot log[\frac{\pi_{\theta}(y \mid x)}{\pi_{ref}(y \mid x)}]\), and the simplified loss of NPO is:
\begin{equation}
L_{NPO} = - \mathbb{E}_{(x,y) \in \mathcal{E}_r} r_{\theta}(y \mid x) + \mathbb{E}_{(x,y) \in \mathcal{E}_f}  r_{\theta}(y \mid x).
\end{equation}
It's obvious that \(L_{NPO} = - \mathbb{D}_{TV}\). Therefore, optimizing the NPO loss essentially means increasing the separation of the edge representations between the retain set and the forget set. 

The gradient of NPO loss can be expressed as follows:
\begin{equation}
\nabla_\theta \mathcal{L}_{\text{NPO}} = -\mathbb{E}_{\mathcal{E}_r} \left[\nabla_{\theta} r_\theta(y \mid x) \right] +\mathbb{E}_{\mathcal{E}_f} \left[\nabla_{\theta} r_\theta(y \mid x) \right].
\end{equation}

When we enhance the influence of edges on the forget set, the revised loss of NPO similar to KL-Divergence is:
\begin{equation}
L_{INPO} = - \mathbb{E}_{(x, y) \in \mathcal{E}_r} r_{\theta}(y \mid x) + \mathbb{E}_{(x, y) \in \mathcal{E}_f} r_{\theta}(y \mid x) \cdot e^{\xi},
\end{equation}
where \(\xi\) is the influence of edges on the forget set. Now the gradient of NPO is:
\begin{equation}
\nabla_\theta \mathcal{L}_{\text{INPO}} = -\mathbb{E}_{\mathcal{E}_r} \left[\nabla_{\theta} r_\theta(y \mid x) \right] +\mathbb{E}_{\mathcal{E}_f} \left[\nabla_{\theta} r_\theta(y \mid x) \cdot e^{\xi} \right].
\end{equation}
The above equation means that giving more attention to the forget set reduces the impact on model utility. Therefore, during the unlearning process, the model reduces the prediction probability of forgetting edges, \textbf{enlarging the probability difference} between the forget set and the retain set to mitigate topological coupling. Simultaneously, it \textbf{preserves the relative influence invariant}.

\subsection{Robust for Different Delete Ratio}
From the comparison in Table \ref{supp:add:ratio:1}-\ref{supp:add:ratio:4}, we can draw the following conclusions:     
\begin{itemize}
    \item \textbf{INPO-S is highly robust to different deletion ratios}. INPO-S consistently outperforms GNNDelet in all metrics of forgetting quality, with \textbf{the MI Ratio being on average \(7.34\%\) higher}. Notably, as the deletion ratio increases, the MI Ratio of INPO-S even improves.
    \item \textbf{True forgetting maintenance}. As the deletion ratio increases, the \( p_f \) of GNNDelete rises with the increase in the deletion ratio, which indicates that true forgetting is not achieved at high deletion ratios. In contrast, the \( p_f \) of INPO-S decreases as the deletion ratio increases, suggesting that \textbf{higher deletion ratios actually promote effective forgetting}. This is the key difference between the two methods.
    \item \textbf{Good performance maintenance}. As the deletion ratio increases, INPO-S maintains the model's AUC and AP on the Retain set just as well as GNNDelete.
\end{itemize}

In summary, INPO-S not only achieves significant improvements in forgetting quality and true forgetting metrics but also demonstrates a remarkable enhancement in robustness against deletion ratios. This illustrates that \textbf{Preference Optimization} is an promising approach for effective graph unlearning.

\begin{table}[h]
\caption{The model utility and forget quality on Cora dataset for INPO-S.}
\label{supp:add:ratio:1}
\vspace{-5pt}
\begin{center}
\begin{small}
\setlength{\tabcolsep}{12pt} 
\begin{tabular}{lcccccc}
\toprule
       & \multicolumn{2}{c}{Retain Set} & \multicolumn{2}{c}{Forget Set} \\
\cmidrule(lr){2-3} \cmidrule(lr){4-5} 
Delete Ratio  & \multicolumn{1}{c}{AUC} & \multicolumn{1}{c}{AP} & \multicolumn{1}{c}{AUC} & \multicolumn{1}{c}{AP} \\
\midrule
0.5         & 0.9640       & 0.9638     & 0.9826       & 0.9857                \\
1.0         & 0.9583       & 0.9575     & 0.9658       & 0.9722               \\
1.5         & 0.9570       & 0.9545     & 0.9546       & 0.9626                \\
2.0         & 0.9559       & 0.9538     & 0.9418       & 0.9514              \\
2.5         & 0.9558       & 0.9534     & 0.9320       & 0.9426                     \\
5.0         & 0.9551       & 0.9532     & 0.9012       & 0.9148                \\
\bottomrule
\end{tabular}
\end{small}
\end{center}
\end{table}

\begin{table}[h]
\caption{The Predicted Probability and MI Ratio on Cora dataset for INPO-S.}
\label{supp:add:ratio:2}
\vspace{-5pt}
\begin{center}
\begin{small}
\setlength{\tabcolsep}{12pt} 
\begin{tabular}{lcccccc}
\toprule
Delete Ratio  & \multicolumn{1}{c}{\(p_f\)} & \multicolumn{1}{c}{\(p_r\)} & \multicolumn{1}{c}{\(\frac{p_r}{p_f}\)} & \multicolumn{1}{c}{MI Ratio} \\
\midrule
0.5         & 0.5473       & 0.8562     & 1.56       & 1.79                \\
1.0         & 0.5555       & 0.8109     & 1.46       & 1.76               \\
1.5         & 0.5526       & 0.7770     & 1.40       & 1.77                \\
2.0         & 0.5514       & 0.7472     & 1.36       & 1.77              \\
2.5         & 0.5507       & 0.7324     & 1.33       & 1.78                    \\
5.0         & 0.5417       & 0.6652     & 1.23       & 1.80                \\
\bottomrule
\end{tabular}
\end{small}
\end{center}
\end{table}

\begin{table}[h]
\caption{The model utility and forget quality on Cora dataset for GNNDelete.}
\label{supp:add:ratio:3}
\vspace{-5pt}
\begin{center}
\begin{small}
\setlength{\tabcolsep}{12pt} 
\begin{tabular}{lcccccc}
\toprule
       & \multicolumn{2}{c}{Retain Set} & \multicolumn{2}{c}{Forget Set} \\
\cmidrule(lr){2-3} \cmidrule(lr){4-5} 
Delete Ratio  & \multicolumn{1}{c}{AUC} & \multicolumn{1}{c}{AP} & \multicolumn{1}{c}{AUC} & \multicolumn{1}{c}{AP} \\
\midrule
0.5         & 0.9609       & 0.9609     & 0.9797       & 0.9838                \\
1.0         & 0.9626       & 0.9619     & 0.9621       & 0.9692               \\
1.5         & 0.9632       & 0.9616     & 0.9487       & 0.9565                \\
2.0         & 0.9643       & 0.9632     & 0.9336       & 0.9412              \\
2.5         & 0.9646       & 0.9634     & 0.9248       & 0.9321                     \\
5.0         & 0.9669       & 0.9673     & 0.8939       & 0.9000               \\
\bottomrule
\end{tabular}
\end{small}
\end{center}
\end{table}

\begin{table}[h]
\caption{The Predicted Probability and MI Ratio on Cora dataset for GNNDelete.}
\label{supp:add:ratio:4}
\vspace{-5pt}
\begin{center}
\begin{small}
\setlength{\tabcolsep}{12pt} 
\begin{tabular}{lcccccc}
\toprule
Delete Ratio  & \multicolumn{1}{c}{\(p_f\)} & \multicolumn{1}{c}{\(p_r\)} & \multicolumn{1}{c}{\(\frac{p_r}{p_f}\)} & \multicolumn{1}{c}{MI Ratio} \\
\midrule
0.5         & 0.5595       & 0.8524     & 1.52       & 1.75                \\
1.0         & 0.5892       & 0.8198     & 1.39       & 1.66               \\
1.5         & 0.5951       & 0.7963     & 1.33       & 1.64                \\
2.0         & 0.5994       & 0.7778     & 1.29       & 1.63              \\
2.5         & 0.6003       & 0.7682     & 1.28       & 1.63                    \\
5.0         & 0.6005       & 0.7297     & 1.21       & 1.63                \\
\bottomrule
\end{tabular}
\end{small}
\end{center}
\end{table}

\subsection{Comparison between GNNDelete and INPO}
\label{appendix:sec:comparison}
An analysis of the principles of GNNDelete reveals the following two significant issues:
\begin{itemize}
    \item \textbf{Limited forgetting capability}. GNNDelete actually distinguishes edges between the retain set and the forget set by introducing \textbf{additional parameters}, which limits its forgetting capability due to the number of parameters. Consequently, as the deletion ratio increases, the quality of forgetting decreases.
    \item \textbf{Structural noise}. The Deleted Edge Consistency(DEC) loss in GNNDelete minimizes the distance between predictions of forget edges and random-chosen node pairs, which would introduce structural noise by encouraging node embeddings in \(\mathcal{E}_f\) to reflect random connections rather than forgetting.
\end{itemize}
The Deleted Edge Consistency loss is:
\begin{equation}
    \mathcal{L}_{\text{DEC}}^l = MSE \Big( \big\{ [h_u^{\prime l}; h_v^{\prime l}] \mid e_{uv} \in \mathcal{E}_f \big\}, \big\{ [h_u^{l}; h_v^{l}] \mid u, v \in_{\mathcal{R}}V \big\} \Big),
\end{equation}
where MSE refers to Mean-Squared Error, and \(\in_{\mathcal{R}}\) means randomly chosen.

We also design experiments from these two perspectives to further validate whether this issue exists in INPO-S as well.
\begin{itemize}
    \item \textbf{Fewer parameters}. We reduce the number of additional parameters to 3/4 and 1/2, and compare the performance of the two models. Implemented using low-rank decomposition.
    \item \textbf{Higher deletion ratio}. We set the deletion ratio to an extremely high value(\(10\%\)) and compare the performance of the two models.
\end{itemize}

\begin{table}[h]
\caption{The model utility and forget quality on Cora dataset for INPO-S with fewer parameters(1/2).}
\label{fig:performance:para:1/2}
\vspace{-5pt}
\begin{center}
\begin{small}
\setlength{\tabcolsep}{12pt} 
\begin{tabular}{lcccccc}
\toprule
       & \multicolumn{2}{c}{Retain Set} & \multicolumn{2}{c}{Forget Set} \\
\cmidrule(lr){2-3} \cmidrule(lr){4-5} 
Model  & \multicolumn{1}{c}{AUC} & \multicolumn{1}{c}{AP} & \multicolumn{1}{c}{AUC} & \multicolumn{1}{c}{AP} \\
\midrule
GNNDelete      & 0.7487       & 0.7664     & 0.7636       & 0.8232                \\
INPO-S         & 0.7950       & 0.8029     & 0.7436       & 0.7902               \\
\bottomrule
\end{tabular}
\end{small}
\end{center}
\end{table}

\begin{table}[h]
\caption{The Predicted Probability and MI Ratio on Cora dataset for INPO-S with fewer parameters(1/2).}
\label{fig:probability:para:1/2}
\vspace{-5pt}
\begin{center}
\begin{small}
\setlength{\tabcolsep}{12pt} 
\begin{tabular}{lcccccc}
\toprule
Model  & \multicolumn{1}{c}{\(p_f\)} & \multicolumn{1}{c}{\(p_r\)} & \multicolumn{1}{c}{\(\frac{p_r}{p_f}\)} & \multicolumn{1}{c}{MI Ratio} \\
\midrule
GNNDelete       & 0.5018       & 0.5531     & 1.10       & 1.95               \\
INPO-S          & 0.5091       & 0.5722     & 1.12       & 1.92               \\
\bottomrule
\end{tabular}
\end{small}
\end{center}
\end{table}

From the comparison of Table \ref{fig:performance:para:1/2} and \ref{fig:probability:para:1/2}, we conclude that \textbf{the capacity of additional parameters also limits the unlearning ability of INPO-S}, meaning that merely designing from the optimization objective cannot resolve this issue. However, compared to GNNDelete, INPO-S can improve the overall AUC and \(\frac{p_r}{p_f}\). This indicates that enlarging the probability gap remains effective even in the case of fewer parameters. The results of Table \ref{fig:performance:para:3/4} and \ref{fig:probability:para:3/4} also confirms this conclusion.

\begin{table}[h]
\caption{The model utility and forget quality on Cora dataset for GNNDelete with fewer parameters(3/4).}
\label{fig:performance:para:3/4}
\vspace{-5pt}
\begin{center}
\begin{small}
\setlength{\tabcolsep}{12pt} 
\begin{tabular}{lcccccc}
\toprule
       & \multicolumn{2}{c}{Retain Set} & \multicolumn{2}{c}{Forget Set} \\
\cmidrule(lr){2-3} \cmidrule(lr){4-5} 
Model  & \multicolumn{1}{c}{AUC} & \multicolumn{1}{c}{AP} & \multicolumn{1}{c}{AUC} & \multicolumn{1}{c}{AP} \\
\midrule
GNNDelete         & 0.7502      & 0.7683      & 0.7696       & 0.8269                \\
INPO-S            & 0.7992      & 0.8085      & 0.7418       & 0.7903              \\
\bottomrule
\end{tabular}
\end{small}
\end{center}
\end{table}

\begin{table}[h]
\caption{The Predicted Probability and MI Ratio on Cora dataset for GNNDelete with fewer parameters(3/4).}
\label{fig:probability:para:3/4}
\vspace{-5pt}
\begin{center}
\begin{small}
\setlength{\tabcolsep}{12pt} 
\begin{tabular}{lcccccc}
\toprule
Model  & \multicolumn{1}{c}{\(p_f\)} & \multicolumn{1}{c}{\(p_r\)} & \multicolumn{1}{c}{\(\frac{p_r}{p_f}\)} & \multicolumn{1}{c}{MI Ratio} \\
\midrule
GNNDelete      & 0.5018       & 0.5544     & 1.10      & 1.95                 \\
INPO-S         & 0.5096       & 0.5770     & 1.13      & 1.92               \\
\bottomrule
\end{tabular}
\end{small}
\end{center}
\end{table}

The results in Tables \ref{fig:performance:del-ratio:10} and \ref{fig:probability:del-ratio:10} show that INPO-S still maintains strong forgetting ability even at very large delete ratio, particularly in the \(\frac{p_r}{p_f}\) metric, outperforming GNNDelete by \(4.95\%\).

\begin{table}[h]
\caption{The model utility and forget quality on Cora dataset for higher deletion ratio(\(10\%\)).}
\label{fig:performance:del-ratio:10}
\vspace{-5pt}
\begin{center}
\begin{small}
\setlength{\tabcolsep}{12pt} 
\begin{tabular}{lcccccc}
\toprule
       & \multicolumn{2}{c}{Retain Set} & \multicolumn{2}{c}{Forget Set} \\
\cmidrule(lr){2-3} \cmidrule(lr){4-5} 
Model  & \multicolumn{1}{c}{AUC} & \multicolumn{1}{c}{AP} & \multicolumn{1}{c}{AUC} & \multicolumn{1}{c}{AP} \\
\midrule
GNNDelete         & 0.9683       & 0.9694     & 0.8685       & 0.8720                \\
INPO-S         & 0.9573       & 0.9559     & 0.8756       & 0.8875               \\
\bottomrule
\end{tabular}
\end{small}
\end{center}
\end{table}

\begin{table}[h]
\caption{The Predicted Probability and MI Ratio on Cora dataset for higher deletion ratio(\(10\%\)).}
\label{fig:probability:del-ratio:10}
\vspace{-5pt}
\begin{center}
\begin{small}
\setlength{\tabcolsep}{12pt} 
\begin{tabular}{lcccccc}
\toprule
Model  & \multicolumn{1}{c}{\(p_f\)} & \multicolumn{1}{c}{\(p_r\)} & \multicolumn{1}{c}{\(\frac{p_r}{p_f}\)} & \multicolumn{1}{c}{MI Ratio} \\
\midrule
GNNDelete         & 0.6041       & 0.6118     & 1.01      & 1.62                \\
INPO-S            & 0.5326       & 0.5639     & 1.06      & 1.84              \\
\bottomrule
\end{tabular}
\end{small}
\end{center}
\end{table}

Additionally, compared to GNNDelete, our INPO model also mitigates the strong coupling between data, and this conclusion remains true. This is why it still performs well even at higher deletion ratios. On the contrary, at extremely high deletion ratios, the impact of structural noise becomes more pronounced, causing \(\frac{p_r}{p_f}\) to approach 1.

\subsection{Mitigate Topological Coupling}
\begin{figure}[t]
  \centering
  \includegraphics[width=1\linewidth]{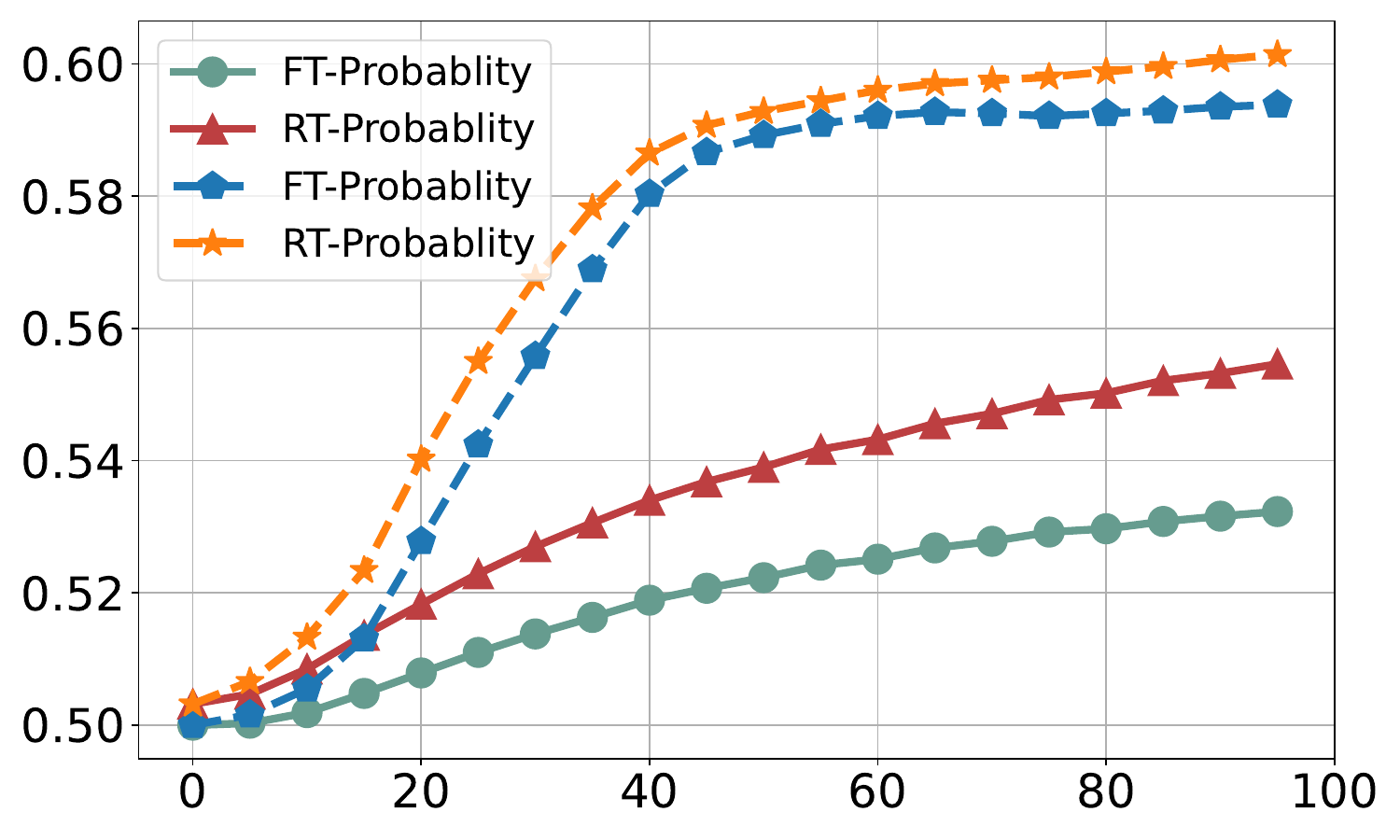}
  \caption{Predicted probability on Cora dataset for delete ratio = \(10\%\). The solid lines denotes INPO-S and the dashed
  lines represents GNNDelete.}
  \label{fig:large:prob:gap}
\end{figure}

INPO enlarges \textbf{the probability difference} between the forget set and the retain set to mitigate the topological coupling. By comparing the \(\frac{p_r}{p_f}\) metric, it is easy to see that INPO achieves effective topological decoupling. Compared to NPO, INPO improves the \(\frac{p_r}{p_f}\) by \(22.22\%\) on DBLP dataset and \(28.16\%\) on Cora dataset. On the other hand, INPO-S achieves a higher \(\frac{p_r}{p_f}\) than GNNDelete at all deletion ratios. Figure \ref{fig:large:prob:gap} shows \textbf{INPO-S achieves large probability difference} and the rate of increase in \(p_r\) is significantly greater than that of \(p_f\).

\begin{figure}[t]
  \centering
  \includegraphics[width=1\linewidth]{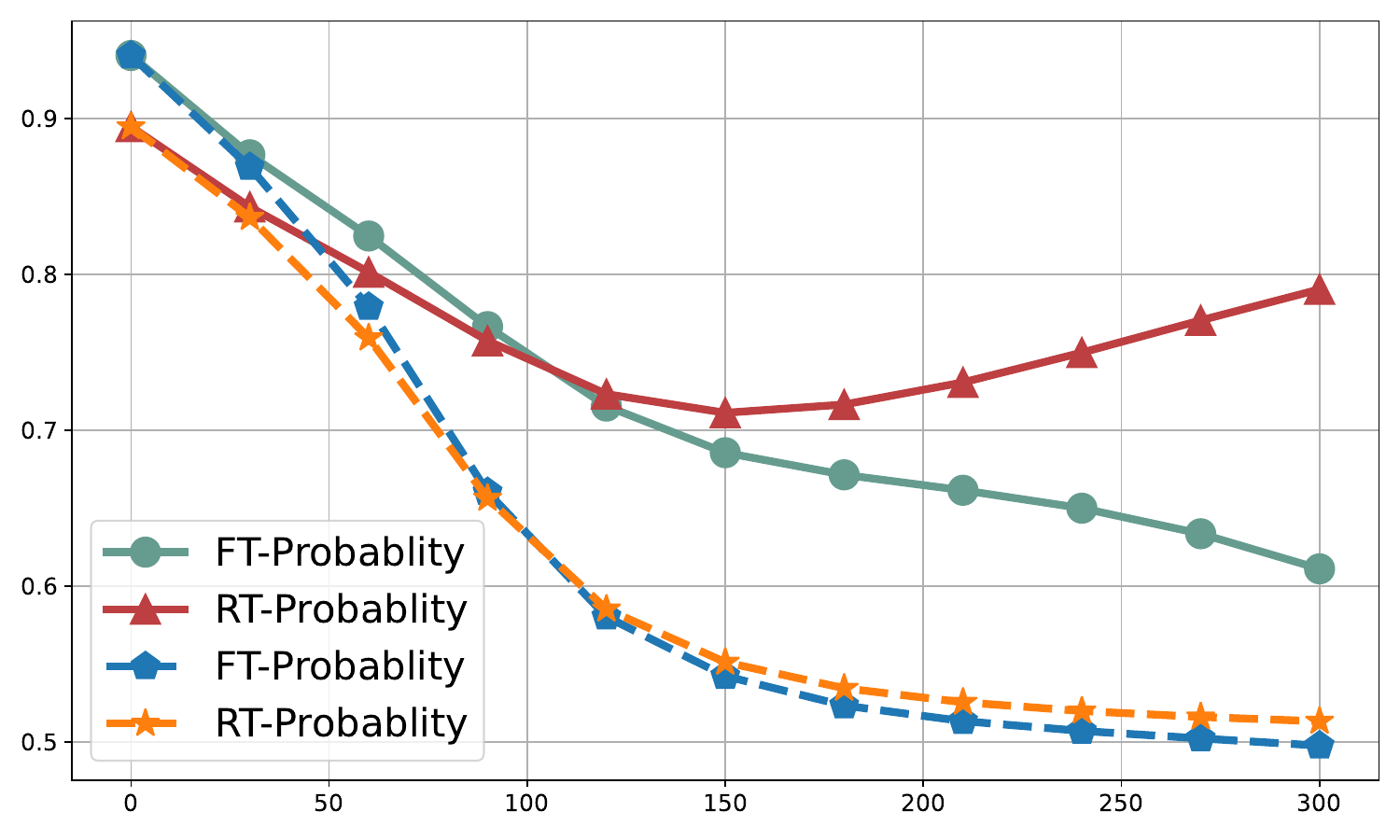}
  \caption{Predicted probability on Cora dataset for delete ratio = \(0.5\%\). The solid lines denotes INPO and the dashed
  lines represents NPO.}
  \label{fig:supp-prob-gap}
\end{figure}

To better illustrate how our method mitigates topological coupling during the training process, we conducted a detailed analysis based on the probability change curves for two scenarios:
\begin{itemize}
    \item \textbf{INPO VS. NPO}. From Figure \ref{fig:supp-prob-gap}, we can draw three conclusions: (1) INPO effectively enlarges the distance between the retain set and the forget set by maintaining the probability \(p_r\). (2) In NPO, \(p_r\) consistently decreases along with \(p_f\) throughout the entire training process. (3) In the early stages of INPO, \(p_r\) also decreases along with \(p_f\), and only later does a turning point occur. This indicates that INPO only becomes effective when \(p_f\) decreases.
    \item \textbf{INPO-S VS. GNNDelete}. INPO-S uses additional parameters to keep \(p_f\) unchanged, thereby achieving unlearning, while continuously improving \(p_r\).
\end{itemize}

In conclusion, in two scenarios, our \textbf{Influence-aware Negative Preference Optimization} achieves effective graph unlearning.

\subsection{Why Do We Need Topological Decoupling?}
Assume that the prediction probability distribution of a unlearned model on the forget set is denoted as \( p_f \), on the retain set as \( p_r \), and on the untrained set (referring to those edges that \textbf{ really do not exist}) as \( p_{ut} \). An ideal unlearned model should satisfy the following two conditions: 
\begin{itemize}
    \item \textbf{In-distinguishability}. \( p_f \) should be as close as possible to \( p_{ut} \);
    \item \textbf{Separability}. \( p_f \) should be as far(separated) as possible from \( p_r \).
\end{itemize}

\begin{figure}[t]
  \centering
  \includegraphics[width=0.8\linewidth]{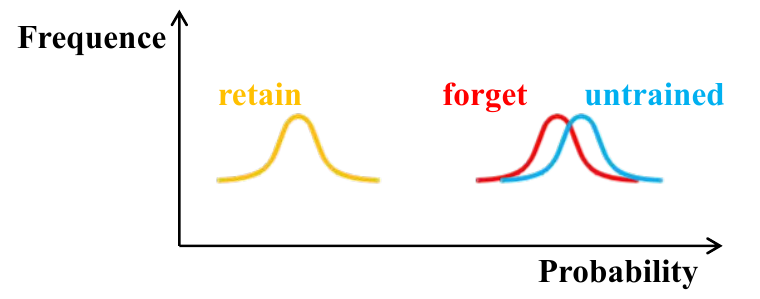}
  \caption{The distribution of ideal unlearned model.}
  \label{fig:ideal}
\end{figure}

\begin{figure}[t]
  \centering
  \includegraphics[width=0.8\linewidth]{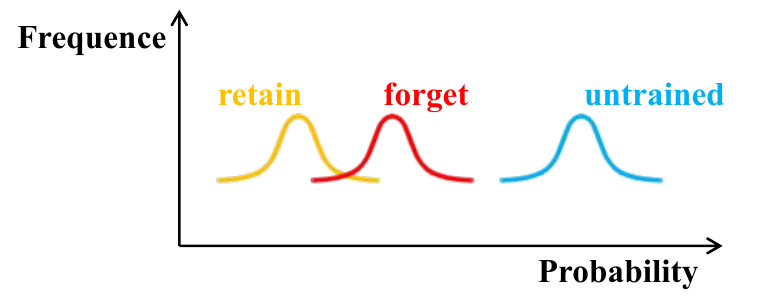}
  \caption{The distribution of current unlearned model.}
  \label{fig:worse:dist}
\end{figure}

We can use Figure \ref{fig:ideal} to represent an ideal unlearned model. However, although most graph unlearning models currently achieve high AUC and AP on the forget set and retain set, they do not perform well in the two aspects mentioned above, such as GNNDelete and MEGU. We list the In-distinguishability and Separability-related data of models in Table \ref{table:distinguishability:separability:Cora}. As can be seen from Table \ref{table:distinguishability:separability:Cora}, directly using NPO for unlearning results in an unlearned model with poor separability (i.e., the unlearning process significantly reduce model utility). This is due to \textbf{the strong coupling between entities in the graph data}, and this result can be illustrated using Figure \ref{fig:worse:dist}.

To address the current challenges faced by these models and NPO, we propose that topological decoupling can better achieve in-distinguishability and separability. Based on Table \ref{table:distinguishability:separability:Cora}, we can draw the following two conclusions: 
\begin{itemize}
    \item Compared with NPO, INPO enlarges the probability difference between the forget set and the retain set to mitigate the topological coupling, achieves higher separability;
    \item INPO-S achieves the best in-distinguishability and separability.
\end{itemize}

\begin{table}[h]
\caption{The In-distinguishability and Separability on Cora. The table shows the mean of the distributions.}
\label{table:distinguishability:separability:Cora}
\vspace{-5pt}
\begin{center}
\begin{small}
\setlength{\tabcolsep}{8pt} 
\begin{tabular}{lcccccc}
\toprule
Model  & \multicolumn{1}{c}{\(p_f\)} & \multicolumn{1}{c}{\(p_{ut}\)} & \multicolumn{1}{c}{\(p_r\)} & \multicolumn{1}{c}{In-dist.} & \multicolumn{1}{c}{Sepa.} \\
\midrule
NPO         & 0.5323       & 0.4726     & 0.5476      & 0.0597  & 0.0153                \\
INPO        & 0.6070       & 0.4812     & 0.8036      & 0.1258  & 0.1966                \\
GNNDelete         & 0.5595       & 0.4968     & 0.8524      & 0.0627   & 0.2929                \\
INPO-S            & 0.5473       & 0.4974     & 0.8562      & 0.0499     & 0.3089         \\
\bottomrule
\end{tabular}
\end{small}
\end{center}
\end{table}

\subsection{Model Implementation Details} 
For NORA, We follow the parameter settings from the original paper, i.e., \(k_1=1.0\), \(k_2=0.5\), \(k_2^{'}=0\), \(k_3=1\), \(\gamma=8\).

In the implementation of the GCN layer, we adopt the most commonly used MPNN framework. Additionally, We use two GCN layers for all graph models. The edge unlearning task is performed on all datasets based on \textbf{the link prediction task}.

In the implementation of INPO-S, We discard the Deleted Edge Consistency in GNNDelete and instead adopt the NPO loss and topological entropy loss to forget specific edges. At the same time, we also follow the idea of \textbf{adding new parameters} in GNNDelete. We add additional parameters after each GCN layer to adjust the representations of the neighbors around the forgotten edges, with the aim of achieving unlearning.

For fewer parameters settings in Sec. \ref{appendix:sec:comparison}, We adopt the approach of \textbf{low-rank decomposition}. The dimension of new additional parameters is \(128\times128\). We decompose it into two low-rank matrices(A, B) to achieve parameter reduction. 
\begin{itemize}
    \item \textbf{1/2 of the total quantity}. The dimension of A is \(128\times32\), and The dimension of B is \(32\times128\);
    \item \textbf{3/4 of the total quantity}. The dimension of A is \(128\times48\), and The dimension of B is \(48\times128\).
\end{itemize}

\subsection{Formal Proof of Lemma 3.1}
The proof proceeds as follows:

1. According to Eq. 2 in main paper, in the unlearning task,  $ \nabla_\theta \mathcal{L}_{\text{NPO}} \ll \nabla_\theta \mathcal{L}_{\text{GA}} $ due to $S_{\theta}(x,y) \ll 1$. Therefore, compared to GA, NPO exhibits very slow parameter updates during unlearning, i.e., \textbf{the divergence speed is slow}. 

2. The update of $\theta$ after unlearning is given by: 
\begin{equation}
 \theta' = \theta - \eta \cdot \nabla_\theta \mathcal{L}_{\text{NPO}}(\theta) .
\end{equation}

3. Considering the change in the loss function on the retain set using a first-order Taylor approximation: 
\begin{equation}
 \mathcal{L}_{\text{retain}}(\theta) = \mathbb{E}_{(x, y) \sim \mathcal{E}_r} \left[ -\log \pi_\theta(y \mid x) \right], 
\end{equation}

\begin{equation}
\begin{aligned}
\mathcal{L}_{\text{retain}}(\theta') &\approx \mathcal{L}_{\text{retain}}(\theta) + \nabla_\theta \mathcal{L}_{\text{retain}}^\top \cdot (\theta' - \theta) \\
&= \mathcal{L}_{\text{retain}}(\theta) - \eta \cdot \nabla_\theta \mathcal{L}_{\text{retain}}^\top \cdot \nabla_\theta \mathcal{L}_{\text{NPO}},
\end{aligned}
\end{equation}
\begin{equation}
 \Delta \mathcal{L}_{\text{retain}} \propto \|\nabla_\theta \mathcal{L}_{\text{NPO}}\| .
\end{equation}
   Therefore, compared to GA, the performance decrease on the retain set occurs more slowly.

\subsection{Formal Proof of Proposition 3.2}
The proof proceeds as follows:

1. We assume $ h^{G}_{\theta}(v_i,v_j) $ is the probability of edge $(v_i, v_j)$ on the original graph computed by the model, and let
\begin{equation}
  G' = G \setminus \{(v_i, v_j)\},
\end{equation}
where the edge $(v_i, v_j)$ is removed from the original graph $ G $. We get the influence of edge $ (v_i, v_j) $:
\begin{equation}
  \xi_{ij} = \| h^{G}_{\theta}(v_i,v_j) - h^{G'}_{\theta}(v_i,v_j) \|.
\end{equation}

2. Given the message passing mechanism of GCN:
\begin{equation}
  H^{(l+1)} = \sigma\left(\hat{A} H^{(l)} W^{(0)}\right),
\end{equation}
and we define the edge probability as
\begin{equation}
  h_\theta(v_i,v_j) = f\left(H_{v_i}^{(L)}, H_{v_j}^{(L)}\right),
\end{equation}
where $ f(\cdot, \cdot) $ is a readout function. Then we get:
\begin{equation}
\begin{aligned}
  \frac{\partial h_\theta(v_i,v_j)}{\partial A_{ij}} 
  &= 
  \frac{\partial f(H_{v_i}^{(L)}, H_{v_j}^{(L)})}{\partial H_{v_i}^{(L)}} 
     \cdot \frac{\partial H_{v_i}^{(L)}}{\partial A_{ij}} 
  + 
  \frac{\partial f(H_{v_i}^{(L)}, H_{v_j}^{(L)})}{\partial H_{v_j}^{(L)}} 
     \cdot \frac{\partial H_{v_j}^{(L)}}{\partial A_{ij}} .
\end{aligned}
\end{equation}

\begin{equation}
\begin{aligned}
  \frac{\partial H_{v_i}^{(L+1)}}{\partial A_{ij}} 
  \propto 
  \frac{\partial \hat{A}_{ij}}{\partial A_{ij}} \cdot H_{v_j}^{(L)} W^{(L)} 
  + 
  \sum_{u \in \mathcal{N}(i)} 
  \frac{\partial \hat{A}_{iu}}{\partial A_{iu}} \cdot H_u^{(L)} W^{(L)} .
\end{aligned}
\end{equation}

3. If node $v_j$ contributes more significantly to the final representation of node $v_i$ (i.e., $H_{v_j}^{(L)}$ has a larger proportion in $H_{v_i}^{(L+1)}$), then $\left\| \frac{\partial H_{v_i}^{(L+1)}}{\partial A_{ij}} \right\|$  is larger.

4. According to Step 3, edges with high influence generate larger $\left\| \frac{\partial H_{v_i}^{(L+1)}}{\partial A_{ij}} \right\|,$
during the optimization process, leading to \textbf{smaller $ \pi_\theta $} by rapid unlearning. Figure 3 in the appendix visualizes this process from the perspective of $S_{\theta}(x,y)$, and values above 0.4 have been reduced. Smaller $S_{\theta}(x,y)$ corresponds to smaller $\nabla_\theta \mathcal{L}_{\text{NPO}}$, i.e., slower divergence speed and reduced impact on the retain set.

The most direct goal of ours is to achieve rapid forgetting, while maintaining the performance on the retain set. This is precisely accomplished by amplifying the influence of the edges to be forgotten.


\end{document}